%% file: main.tex
\documentclass[twoside,11pt]{article}
\usepackage{jair, theapa, rawfonts}

\usepackage{times}
\usepackage{helvet}
\usepackage{courier}
\usepackage{graphicx}
\usepackage{xifthen}
\usepackage{fancyvrb}
\usepackage{xcolor}
\usepackage{url}
\usepackage[textsize=scriptsize,textwidth=1cm]{todonotes}
\usepackage{amsmath,amsfonts,amssymb,amsthm}
\usepackage{algorithm}
\usepackage[noend]{algpseudocode}
\usepackage{xpatch}
\usepackage{multirow}
\usepackage{booktabs}

\usepackage{fancyvrb}

\newcommand{\mytitle}{Planning for Goal-Oriented Dialogue Systems}

\jairheading{1}{\textit{under review}}{1-42}{10/19}{??}
\ShortHeadings{\mytitle}{Muise, Chakraborti, Agarwal, Bajgar, Chaudhary, Lastras, Ondrej, Vodol\'an, \& Wiecha}
\firstpageno{1}

\include{macros}

\begin{document}

% \tableofcontents
% \newpage

\title{\mytitle}

\author{
    \name Christian Muise \email christian.muise@ibm.com \\
    \addr IBM Research AI, Cambridge, USA \\
    \AND
    \name Tathagata Chakraborti \email tchakra2@ibm.com \\
    \addr IBM Research AI, Cambridge, USA \\
    \AND
    \name Shubham Agarwal \email Shubham.Agarwal@ibm.com \\
    \addr IBM Research AI, Cambridge, USA \\
    \AND
    \name Ondrej Bajgar\thanks{Work done when at IBM Watson} \email ondrej@bajgar.org \\
    \addr Future of Humanity Institute, University of Oxford, UK \\
    \AND
    \name Arunima Chaudhary \email arunima.chaudhary@ibm.com \\
    \addr IBM Research AI, Cambridge, USA \\
    \AND
    \name Luis A. Lastras-Monta\~no \email lastrasl@us.ibm.com \\
    \addr IBM Research AI, Yorktown Heights, USA \\
    \AND
    \name Josef Ondrej \email josef.ondrej@ibm.com \\
    \addr IBM Watson, Praha, Czech Republic \\
    \AND
    \name Miroslav Vodol\'an \email MVodolan@cz.ibm.com \\
    \addr IBM Watson, Praha, Czech Republic \\
    \AND
    \name Charlie Wiecha \email wiecha@us.ibm.com \\
    \addr Watson Data and AI, Yorktown Heights, USA
}
    
    % \addr IBM Research, Cambridge, USA \\
    % \AND
    % \addr IBM Research, Yorktown Heights, USA \\
    % \AND
    % \addr Future of Humanity Institute, Oxford, UK \\
    % \AND
    % \addr IBM Watson, Praha, Czech Republic \\
    % \AND

    \maketitle

    \input{sections/abstract}

\input{sections/introduction}

\input{sections/framework}

\input{sections/background}

\input{sections/model}

\input{sections/mai}

\input{sections/executor}

    \input{sections/hovor-frontend}
    \input{sections/evaluation}

\input{sections/related}

\input{sections/discussion}

\input{sections/acknowledgements}
    
    % \vskip 0.2in
    \bibliography{references}
    \bibliographystyle{theapa}

\end{document}

%% file: macros.tex
\newcommand{\optarg}[1]{\ifthenelse{\isempty{#1}}{}{(#1)}}

\newcommand{\hide}[1]{}

\newcommand{\la}{\langle}
\newcommand{\ra}{\rangle}

\newcommand{\set}[1]{\lbrace #1 \rbrace}
\newcommand{\st}{\hspace{1mm}|\hspace{1mm}}

\newcommand{\fluents}{\mathcal{F}}
\newcommand{\actions}{\mathcal{A}}
\newcommand{\init}{\mathcal{I}}
\newcommand{\goal}{\mathcal{G}}

\newcommand{\nodes}{\mathcal{N}}
\newcommand{\edges}{\mathcal{E}}
\newcommand{\initnode}{n_0}

\newcommand{\problem}{\la \fluents, \init, \actions, \goal \ra}
\newcommand{\solution}{\la \nodes, \edges, \initnode \ra}

\newcommand{\pre}[1]{\mathrm{Pre}_{#1}}
\newcommand{\effs}[1]{\mathrm{Eff}_{#1}}

\newcommand{\add}[1]{ADD_{#1}}
\newcommand{\del}[1]{DEL_{#1}}

\newcommand{\effand}[1]{\bigwedge_{#1}}
\newcommand{\effoneof}[1]{\bigoplus_{#1}}

\newcommand{\exec}[1]{AEF_{#1}}
\newcommand{\determiner}[1]{DET_{#1}}

\newcommand{\actmap}[1]{action\optarg{#1}}
\newcommand{\edgemap}[2]{next_{#1}\optarg{#2}}
\newcommand{\outgoing}[1]{successors(#1)}
\newcommand{\realizations}[1]{\mathcal{R}\optarg{#1}}

\newcommand{\hovor}{\texttt{Hovor}}
\newcommand{\mai}{\texttt{D3WA}}
\newcommand{\cdd}{\texttt{CDD}}
\newcommand{\dba}{\texttt{D3BA}}

\theoremstyle{definition}

\newtheorem{definition}{Definition}

%% file: sections/abstract.tex
\begin{abstract}
Generating complex multi-turn goal oriented dialogue agents is a difficult problem that has seen considerable focus from many leaders in the tech industry, including IBM, Google, Amazon, and Microsoft. This is in large part due to the rapidly growing market demand for dialogue agents capable of goal-oriented behaviour. Due to the business process nature of these conversations, end-to-end machine learning systems are generally not a viable option, as the generated dialogue agents must be deployable and verifiable on behalf of the businesses authoring them.

In this work, we propose a paradigm shift in the creation of goal-oriented complex dialogue systems that dramatically eliminates the need for a designer to manually specify a dialogue tree, which nearly all current systems have to resort to when the interaction pattern falls outside standard patterns such as slot filling. We propose a \emph{declarative} representation of the dialogue agent to be processed by state-of-the-art planning technology.
Our proposed approach covers all aspects of the process; from model solicitation to the execution of the generated plans / dialogue agents. Along the way, we introduce novel planning encodings for declarative dialogue synthesis, a variety of interfaces for working with the specification as a dialogue architect, and a robust executor for generalized contingent plans. We have created prototype implementations of all components, and in this paper we further demonstrate the resulting system empirically.
\end{abstract}

%% file: sections/introduction.tex
\section{Introduction}

Generating useful dialogue agents that can handle complex tasks is both a grand challenge for the academic field and a major focus of industry.
The market for dialogue agents is estimated to reach \$1.23 billion by 2025 \cite{chatbotreport}. This is a high-growth application area for artificial intelligence techniques, and one in which principled and hybrid approaches that mix symbolic and learning techniques can have a large impact. In this work, we focus on a specific flavour of dialogue agents that is common for business applications: \emph{those that are goal-oriented and multi-turn in nature}. 
A common example of this is in customer support agents, whose goal
is to figure out and solve the customer's problem through conversation.
Viewed through the lens of automated planning, and non-deterministic planning in particular, we tackle the problem of generating these complex agents in a drastically different way than the industry norm, which is to either exhaustively enumerate the space of dialogue flows by hand or to offer special purpose solutions for narrower tasks, such as information gathering (e.g., through the use of slot filling).

Common frameworks for the creation of dialogue agents
in enterprise applications \cite{kinaSUR} 
include, among others, IBM's Watson Assistant\footnote{Watson Assistant: \url{https://www.ibm.com/cloud/watson-assistant/}}, 
Google's Dialogflow\footnote{Dialogflow: \url{https://cloud.google.com/dialogflow/}},
Amazon's Lex\footnote{Amazon Lex: \url{https://aws.amazon.com/lex/}} and Microsoft's Bot Framework \footnote{Microsoft Bot Framework: \url{https://dev.botframework.com/}}. These frameworks all recognize the difficulty of writing and maintaining hand built dialogue trees and have partially addressed this problem by implementing slot filling dialogue modules in which a complex implied graph is traversed with the goal of gathering a given set of slots from the end user. Unfortunately, if the desired interaction deviates from this pattern, these frameworks suffer from a similar drawback: the variety of conversation paths must be explicitly captured in complex and hard-to-maintain dialogue trees or graphs. One possibility which has received significant attention is to use end-to-end trained machine learning architectures, which have the allure of obviating the need to specify any state machine. 

The initial proposals in this direction were based on sequence-to-sequence architectures employing recurrent neural networks  \cite{tay}, which are now understood to be suitable mainly for chit-chat applications in which there is no apparent goal to a conversation and where there is no need to integrate system actions or even give guarantees of behavior. Subsequent proposals focused on variants of the slot-filling paradigm, explicitly calling out a state space that needs to be tracked, API calls that need to be made, and measuring the quality of the language being produced on a turn-basis. It is likely fair to state that these types of techniques are very much still in their research infancy; not only is it generally difficult to provide insight into why a particular dialogue agent answer is being provided (let alone how to change its behavior in a reliable way), it is a difficult problem to create and maintain training data for such systems.

%Figure \ref{fig:car-inspection-log} shows one such conversation unfold.

\begin{figure}[!t]
\centering
\includegraphics[width=0.9\linewidth]{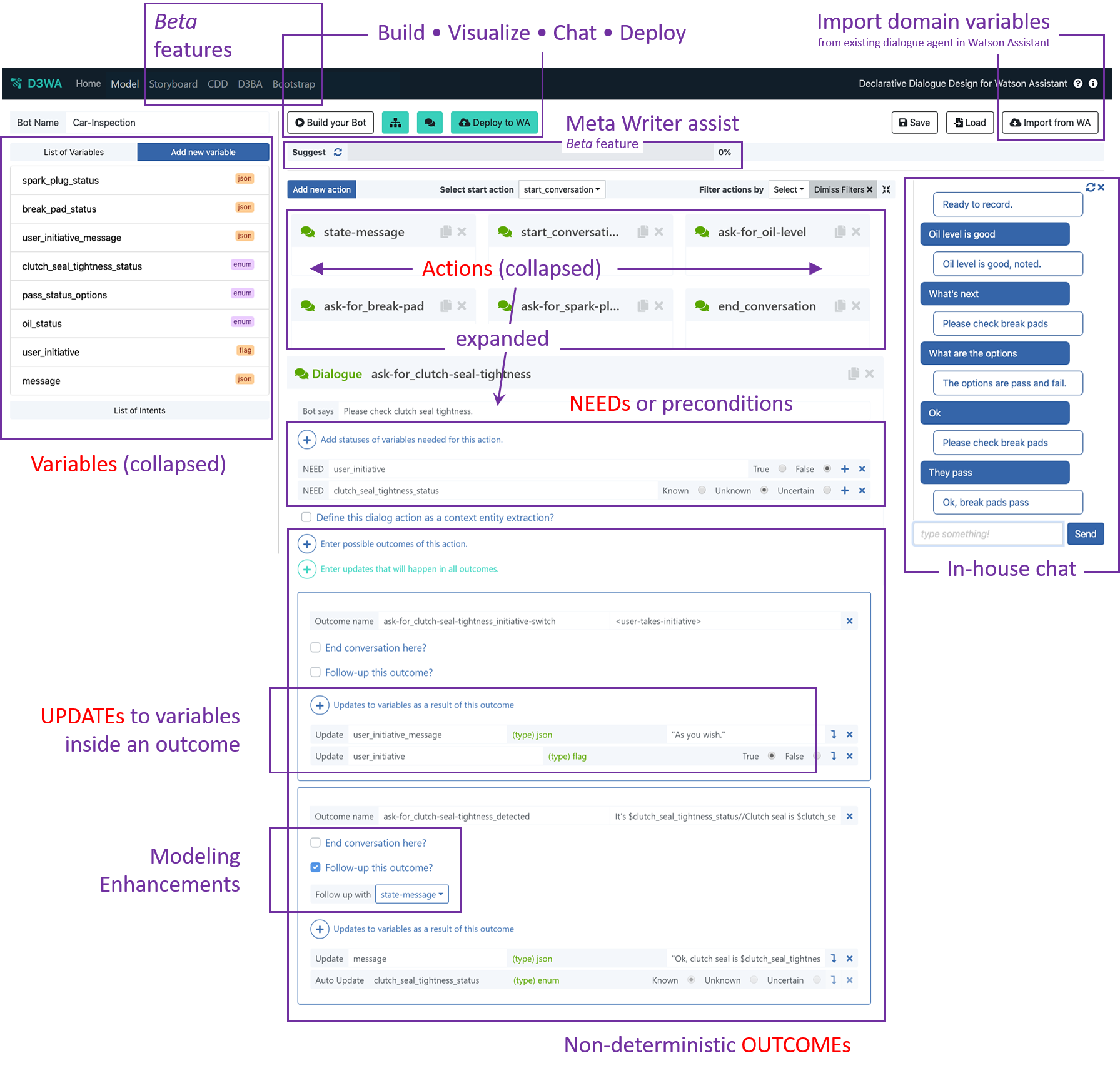}
\caption{Illustration of our interface for declarative dialogue design. The individual components highlighted here are described in detail in Section~\ref{sec:mai}, while
the {\em beta} features are described in Section~\ref{sec:future}.
The visualize button opens into an in-house visualizer 
replacing the action cards in this image, as shown in 
Figure~\ref{fig:generated-plan-inspection}.
}
\label{fig:d3wa-inspection}
\end{figure}

\begin{figure}[!t]
\centering
\includegraphics[width=\columnwidth]{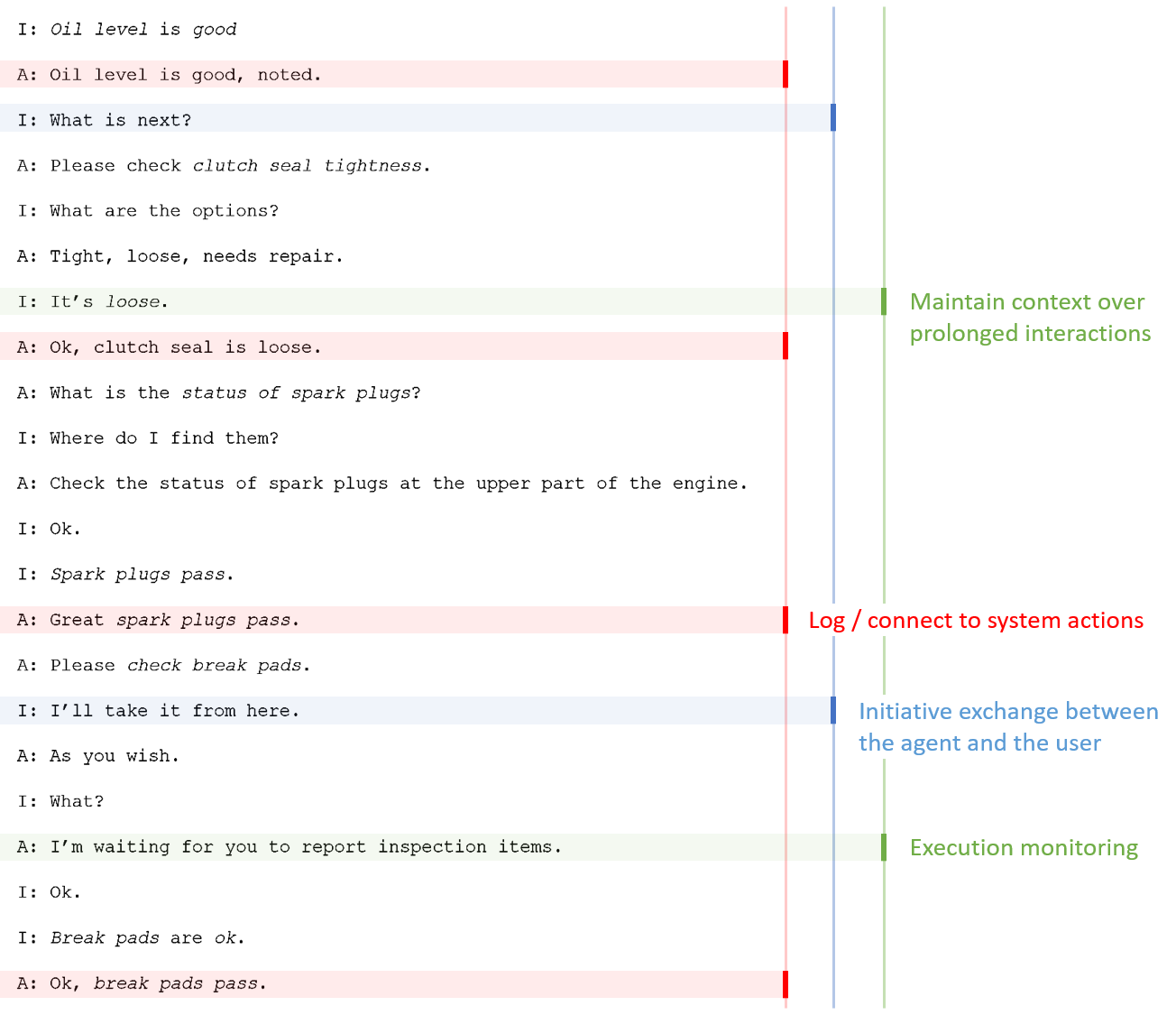}
\caption{Log of a typical multi-turn conversation in the Car Inspection
domain where an operator and the AI agent interactively inspect various
components of a car. The conversation involves complex
initiative switches where the operator or the agent takes control 
of the process. The example also highlights differences with chitchat forms
of dialogue, since the outcome of the inspection needs to be logged 
in practice by connecting to actual system actions in the backend, and sophisticated understanding of context needs to be maintained
and monitored over prolonged interactions with the user.}
\label{fig:car-inspection-log}
\end{figure}

\begin{figure}[!t]
\centering
\includegraphics[width=\linewidth]{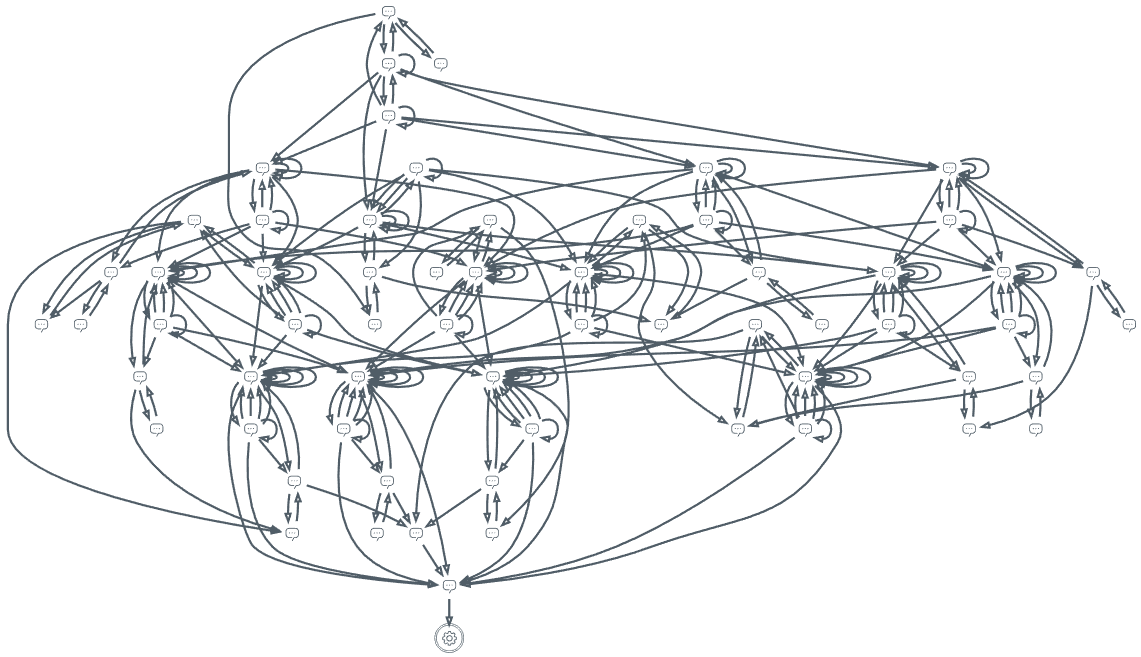}
\caption{Generated dialogue agent for the car inspection domain in Figure
\ref{fig:car-inspection-log}. The declarative specification in view above 
with 7 actions and 8 variables gives rise to this sophisticated dialogue
agent with 63 nodes and 272 edges.}
\label{fig:generated-plan-inspection}
\end{figure}

By recognizing that many aspects of the complex dialogue agents are in fact similar or identical snippets of an underlying process, we arrive at a paradigm shift in how the dialogue agents are specified: \textit{rather than explicitly creating and maintaining entire dialogue graphs, we specify the behaviour declaratively and compile the complete implicit graphs from this compact specification}.
Figure \ref{fig:d3wa-inspection} provides a glimpse of the modeling interface we have created for dialogue designers (to be detailed later in Section \ref{sec:mai}), and it shows an agent created for the car inspection domain introduced above in Figure~\ref{fig:car-inspection-log} with only 7 dialogue actions defined. The ways in which these actions can interact and be orchestrated leads to a dialogue agent where the complexity can be immediately seen with 63 nodes and 272 edges (Figure \ref{fig:generated-plan-inspection}). Creating such agents explicitly by hand is not out of the realm of possibility, but the improved scalability at the declarative level is immediately apparent.

At a high level, we use non-deterministic planning technology to compile declarative specifications of dialogue agents into complete dialogue plans. However, our contributions go far beyond just defining a declarative representation of dialogue design: we have implemented a complete framework that covers the gambit of model acquisition from subject matter experts and dialogue designers (who, notably, have little or no experience with planning technology); all the way to a deployable system capable of communicating with end users and implementing verifiable processes as server API calls.
In this paper, we outline all of these components as well as their integration.\footnote{Note that some sub-components of our work have been published as individual workshop papers \cite{fss} and demos \cite{mai}, but this paper represents the first holistic view of the project.}

% We begin by detailing the general framework in the next section, and providing some preliminary background material in Section \ref{sec:background}. In Section \ref{sec:model} we provide details on the planning model that we use, and the encoding strategies that were employed as a result. In Section \ref{sec:mai} we explore the model acquisition interface, in Section \ref{sec:exec} the plan executor, and in Section \ref{sec:hovor-frontend} the diagnostic frontend interface to the executor. We provide some empirical evaluations of various aspects of our approach in \ref{sec:eval}, and conclude with a discussion of related work and future extensions in Sections \ref{sec:related}-\ref{sec:future}.

%% file: sections/framework.tex
\section{General Framework and Paper Outline}
\label{sec:framework}

\begin{figure}
\centering
\includegraphics[width=\columnwidth]{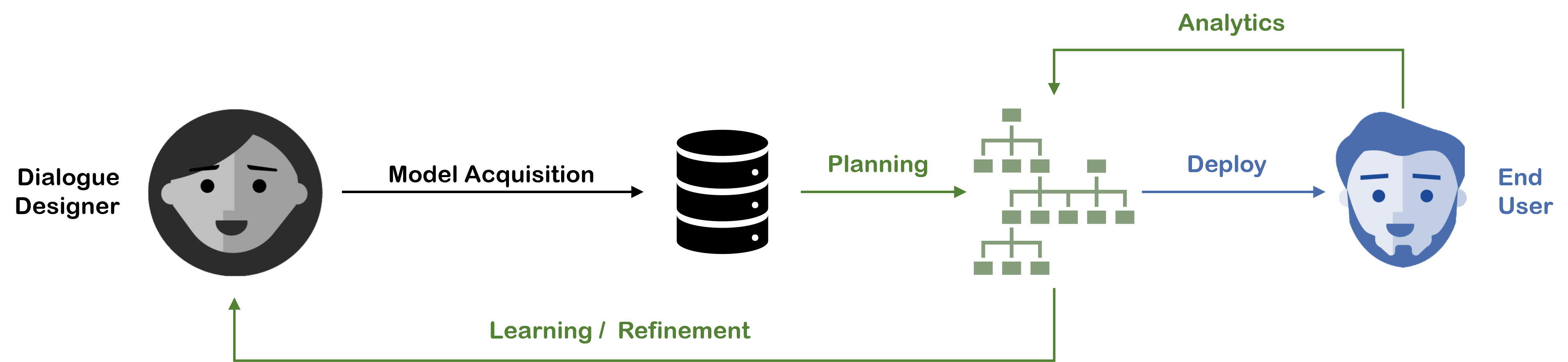}
\caption{Illustration of the different stages of building, deploying and maintaining 
dialogue agents. Our proposed system works end to end across
this development pipeline.}
\label{fig:overview}
\end{figure}

The general architecture of our approach is presented in Figure \ref{fig:overview}. Starting from left, a dialogue designer, working with subject matter experts and developers, models a dialogue agent using an interface that allows for a compact declarative specification -- this phase is described in Section \ref{sec:mai}. The configuration from this interface produces two artefacts: (1) a planning model that represents an abstraction of the problem suitable for non-deterministic solvers to synthesize a complete contingent controller (i.e., the skeleton of our dialogue plan); and (2) a complete configuration required for execution of the plan (including natural language utterances, API endpoint payloads, etc.).

Hidden to the end-user, the planning abstraction (represented using a variant of the Planning Domain Definition Language (PDDL) \cite{pddlbook}) is sent to a remote solver for the synthesis of a plan, and the resulting plan is combined with the remaining configuration to be stored as a deployable dialogue agent. The model abstraction and encoding strategies we use is described in detail in Section \ref{sec:model}.

On right of the diagram, we have the execution of the dialogue agent and interaction with the end user -- the person who chats with the agent. 
Often overlooked in the planning literature, the \textit{effective execution} of these generated plans is paramount to the approach. We describe the custom executor, and the key challenges it overcomes, in Section \ref{sec:exec}.

Finally, the diagram shows the diagnostic view of the deployed agent, and how the dialogue designer can leverage usage information to iteratively improve on the agent's design. We present the interface covering this aspect of the work in Section \ref{sec:hovor-frontend}. 
We will provide empirical evaluations of various stages of our approach in \ref{sec:eval}, and conclude with a discussion of related work and future extensions in Sections \ref{sec:related}-\ref{sec:summary}.
% We have only scratched the surface of what is possible in this regard, but touch on some of the ideas we are exploring next in Section \ref{sec:future}.

%% file: sections/background.tex
\section{Background}
\label{sec:background}

\subsection{Planning}
\label{ssec:planning}

The style of plans we consider are contingent plans that would be generated from a Fully Observable Non-Deterministic (FOND) or Fully Observable Probabilistic (FOP) planner \cite{muise-icaps12-long,camacho-icaps16-probprp}. We do not go into details on how these plans are produced, but instead focus on the planning problem and solution specifications (the latter being a core component of the input for execution). Unlike traditional planning specifications, we assume additional information is available to specify how actions are executed and observed (using callback functions described below).

Here, we review some of the commonly used notation, and extend it to the rich setting required for effective execution.\footnote{For those familiar with non-deterministic formalisms, we do assume fairness of the domain model with some caveats. We discuss the implications of this assumption in Section \ref{sec:model}.} For simplicity, we will assume a FOND setting, but all of the techniques work equally well in the probabilistic setting as well: the key difference between the FOND and probabilistic setting lays with how the plans are produced, as the solution form is the same. The execution of a probabilistic plan need not take into account the probabilities assigned to action effects, and so we do not consider it further.

\begin{definition}[Planning Problem]
A FOND planning problem $\problem$ consists of fluents $\fluents$ that describe what is true or false in the world, an initial state $\init \subseteq \fluents$ where the planning agent begins execution, a set of actions $\actions$ that the agent can do, and a goal $\goal \subseteq \fluents$ that specifies the partial assignment of fluents that must be achieved. A \emph{complete state} (or just \emph{state}) is a subset of the fluents $\fluents$ that are presumed to be true (all other presumed to be false); a \emph{partial state} is similarly defined but without any presumption about the truth of fluents outside of the set.
\end{definition}

Generally, we adopt the standard notation for FOND planning. The one exception is that we do \textit{not} make the simplifying assumption that the action effects are just a set of one or more non-deterministic effects, one of which will be chosen during execution, as is usually assumed in FOND planning. In practice, the action effects are a nesting of \textit{and} and \textit{oneof} clauses (the latter referring to the notion that \textit{exactly one} of the sub-clauses must be used).

We retain this complexity for two key reasons: (1) the arbitrary nesting gives us a level of sophistication that the plan executor can tap into (demonstrated in Section \ref{ssec:complex_outcome_determination} and the evaluation later); and (2) it provides a far more natural form of action specification for the modeler to work with during design and maintenance of the planning model.

\begin{definition}[Complex Actions]
Let $a \in \actions$ be an action. $\pre{a} \subseteq \fluents$ then denotes the \emph{precondition} of $a$  -- the set of fluents that must hold for $a$ to be applicable (that means $a$ is applicable in state $s$ iff $\pre{a} \subseteq s$). $\effs{a}$ denotes the effect formula of $a$. An effect (sub)formula is one of the following:

\vspace{0.5em}
\begin{tabular}{rl}
    $(\neg)f$: & a fluent $f \in \fluents$ or its negation\\[0.5em]
    $\effand{\varphi}$: & a conjunction of effect subformulae \\[0.5em]
    $\effoneof{\varphi}$: & a mutually exclusive disjunction of effect subformulae (i.e., exactly one is chosen)
\end{tabular}
\vspace{0.5em}

\noindent The effect of an action can be viewed as an \textit{and-or} tree. A \emph{realization} of the effect consists of all fluents or their negations that appear in the sub-tree which includes exactly one child of each \textit{or} node (here, we refer to them as \textit{oneof} nodes, and their children as \textit{outcomes}) and all children of each \textit{and} node. Such a realization can be thought of as one possible result of the action's execution. We use $\realizations{a}$ to refer to the set of all realizations for action $a$, and for each realization $r \in \realizations{a}$, we define $\del{r}$ to be the set of fluents removed from the state as a result of this action and $\add{r}$ the fluents to be added. We make the non-standard (but non-restricting) assumption that for all actions and their realizations, $\add{r} \cap \del{r} = \emptyset$.
\end{definition}

The arbitrary nesting of fluents, $\effand{}$, and $\effoneof{}$ operators mirror the common description of FOND problems in PDDL using \texttt{and} and \texttt{oneof} clauses \cite{DBLP:series/synthesis/2013Geffner}. For execution purposes, we assume that every sub-formula of an action's effect is uniquely identifiable, and will use set notation to refer to those components (e.g., $\varphi \in \effs{a}$). Note that we only allow negation at the leaf level of the effect, rather than allow arbitrary negation of sub-formulae. This is due to the fact that the negation of $\effoneof{}$ clauses is ill-defined from the planning perspective, as is the use of $or$ in effects, which would arise from negating an \textit{and} clause (this is in contrast with the arbitrary nesting typically allowed in action preconditions in the literature).

Solutions to a FOND problem generally come in two flavours: (1) policies mapping each state of the world to an action, and (2) contingent plans represented as controllers, where the nodes and edges respectively correspond to actions and possible outcomes \cite{DBLP:series/synthesis/2013Geffner}. We adopt the latter in this work.

\begin{definition}[Contingent Plan]
A solution to a FOND planning problem (or \textit{contingent plan}) is a graph $\solution$, where $\nodes$ are the nodes of the graph corresponding to the actions the agent should take and $\edges$ are the edges corresponding to the possible outcomes of each action associated with the nodes. We use $\initnode \in \nodes$ to refer to the initial node in which the agent should begin executing. We further assume that we have a function mapping nodes to actions ($\actmap{}: \nodes \rightarrow \actions$) and functions mapping the realizations to the successors of a node ($\edgemap{n}{}: \realizations{\actmap{n}} \rightarrow \outgoing{n}$).
\end{definition}

The extra notation for mapping nodes and edges to the original FOND problem allows us to tie together the generated plan and its execution. We make no assumptions on the embodiment of the executing agent, but assume that black-box callback functions are available for (1) the initial execution of the action (which affects the world) and (2) the realization of that action's impact.

\begin{definition}[Callback Functions]
We define the \textit{action execution function} $\exec{a}$ to be the function used to execute planning action $a \in \actions$. We define the \textit{determiner} $\determiner{o}$ to be the function that is used to determine which outcome of the non-deterministic effect $o = \effoneof{\varphi} \varphi$ has occurred. These are blackbox implementations that have certain properties we describe later.
\end{definition}

\subsection{Dialogue Systems \& Natural Language Understanding (NLU)}
\label{sec:background-nl}

We do not require any of the formal constructs from the NLU literature to present our work. However, to provide context for the implementation of our methods, in this section we present some preliminary details on the high-level NLU concepts.

The vast majority of the frameworks for building dialog systems extract \emph{intents} and \emph{entities} from individual end user utterances in order to determine the appropriate course of action. We assume the same natural language understanding model largely because it is a convenient assumption as then the resulting system can inter-operate easily with existing dialogue systems. This type of natural language understanding is relatively coarse and does not take advantage of many developments in the natural language understanding community, such as Abstract Meaning Representation (AMR). The intents and entities model is used due to the fact that it is relatively easy to understand by subject matter experts who otherwise have relatively little machine learning/information extraction skills. It should be noted that our work can easily be extended to handle scenarios where more complex sentence understanding techniques are being used, since the most fundamental assumption that we make is that action preconditions and outcome specifications can be written as logical assertions on variables extracted from user utterances and API results.

In the intent and entities model, typically the subject matter expert gives examples of how an end user may specify a given intent (for example ``I would like to book a flight''). Entities target specific parts of a sentence; for example from the utterance ``I would like to depart tomorrow'' the system is able to extract a date (calculated to be the day after today) or from an utterance ``I would prefer an aisle seat'' an entity can be used to deduce the type of seat being requested. Our framework supports system entities provided by default by IBM Watson Assistant: numbers, percentages, currency, dates, times, locations, and names of people. In addition it supports custom entities defined by the end user as also allowed by Watson Assistant. Finally, our framework also supports defining entities within a context in which they are being used (e.g. ``I want a \textbf{vegetarian} entree and please only \textbf{window} seats'' where \textbf{vegetarian} and \textbf{window} are annotated to be \texttt{\$food\_type} and \texttt{\$seat\_type}, respectively), using the corresponding technology in Watson Assistant.

%% file: sections/model.tex
\section{A Planning Model for Dialogue}
\label{sec:model}

When it comes to orchestrating the sequence of elements that make up a dialogue agent (including the exact utterances / questions asked, endpoints called, etc), there is a lot of extraneous information that can (and should) be abstracted away. In this section, we detail the precise abstraction that is used to model the synthesis of dialogue agents as fully-observable non-deterministic planning, and detail the PDDL encoding strategies that were employed to make the modeling process smoother.

\subsection{Core Encoding}
\label{sec:core-encoding}
Before going into the finer details of the PDDL encoding currently used, we will first describe the key elements of the base encoding. The core PDDL encoding of our declarative dialogue agents consists of (1) an abstraction of the environment variables; (2) specification of three action types (dialogue, system, and web); and (3) a goal specification.

The latter is the simplest to describe: the dialogue designer can flag any outcome of any action to be a ``goal-achieving outcome''. The ramification of this is that an auxiliary \verb|Goal| fluent is added as an effect of that outcome, and the goal for \textit{every} dialogue agent is the same: to achieve \verb|Goal|.

For part (1) (i.e., the state abstraction), we will defer most of the details to Section \ref{sec:exec}, as it is the executor component that must deal with aligning real-world values with the state abstraction the planner uses. The general idea, however, is that complex and arbitrary variables (such as \verb|user_location|) are modeled using a knowledge-based encoding (e.g., with a fluent \verb|(have_user_location)|). For more complex conditions and fluents, individual ``flags'' may be defined, and maintained through common action-theory encoding strategies (e.g., an action with a complex precondition that has a single effect of setting a flag to true).

Finally, we have the three following action types that are generated as part of the compilation.

\subsubsection*{Dialogue Action}
A dialogue action corresponds to a single utterance that the agent sends to the end-user. The non-deterministic outcomes of a dialogue action correspond to how the end-user might respond. As an example, an action \textsc{AskDestination} may correspond to sending the user the message ``Where are you going?'', and the outcomes correspond to (a) the user providing a valid location; (b) the user declaring they no longer want to take a trip; or (c) the user responding in a way that is unrecognized. This is an example PDDL snippet for such an action:

\begin{Verbatim}[frame=single,framesep=5mm,commandchars=\\\{\}]
{\color{gray}; E.g., "Where would you like to go?}
(:{\color{blue}action} ask-destination

    :{\color{blue}precondition} (\textbf{and} (need-destination)
                       (\textbf{not} (have-destination)))

            {\color{gray}; Oneof clauses and outcomes are labeled}
    :{\color{blue}effect} (\textbf{labeled-oneof} {\color{olive}resolve-location-response}
    
                {\color{gray}; E.g., "Whistler, BC."}
                (\textbf{outcome} {\color{olive}extracted-destination}
                  (have-destination))

                {\color{gray}; E.g., "Nowhere now"}
                (\textbf{outcome} {\color{olive}canceled-trip}
                  (\textbf{not} (need-destination)))
                
                {\color{gray}; E.g., "fhqwhgads"}
                (\textbf{outcome} {\color{olive}fallback}
                  (must-clarify))))
\end{Verbatim}

There are a few important things worth noting here. First, we do not have individual outcomes for all of the different locations the user may respond with: the (seemingly) unbounded possibilities are collapsed into a single outcome. This is key, as our abstraction only needs to distinguish between the outcomes listed above, and not the actual realization of the outcome. Of course, the user's ultimate destination must be stored, but this is the task of the plan executor (cf. Section \ref{sec:exec}), and not appropriate for the planning abstraction.

Second, note that all elements of natural language processing (either the generation of the agent utterance or the understanding of the user's response) is absent from the PDDL (the only inclusion of actual utterance text is in the PDDL comments). Again, this is intentional, as the abstraction need not be tied to the grounded dialogue.

Finally, while we \textit{do} extend the traditional PDDL syntax slightly to annotate the action outcomes and oneof clauses (indicated in {\color{olive}olive} in the example above), which aids in execution of the computed plan, there is no direct tie-in to how the outcome will be chosen. Ultimately, this is again an aspect of the plan executor, and in particular for dialogue actions, this will be done using intent recognition with each outcome having a unique intent associated with it. This also leads us to the common practice of having a fallback intent for every dialogue action, as there is always a chance that the NLU will fail to understand what the user has said.

To keep the PDDL examples concise, for the remainder of the text we will forgo showing the \textit{labeled} version of the \textbf{oneof} clauses and outcomes in lieu of the standard syntax.

\subsubsection*{Web Action}
The second type of action we have modeled is a \textit{web action}: the execution of which corresponds to making an endpoint call with a certain JSON payload, and the outcomes of which correspond to the ways the server might respond. Similar to the vast array of end-user responses to a dialogue action, there is a vast array of web server responses we might encounter (e.g., all of the HTTP status code specifications). We again appeal to the right level of abstraction: if all error codes for a web service should be handled similarly (e.g., by telling the user that the server is down, and executing a contingency action), then these would all correspond to a single outcome. Here is an example snippet of the PDDL for a web action:

\begin{Verbatim}[frame=single,framesep=5mm,commandchars=\\\{\}]
{\color{gray}; API call to hotel service}
(:{\color{blue}action} check-availability

    :{\color{blue}precondition} (\textbf{and} (have-destination)
                       (have-travel-dates))

    :{\color{blue}effect} (\textbf{oneof}
                {\color{gray}; Dates work fine}
                (travel-ok)

                {\color{gray}; Dates don't work}
                (must-reschedule)
                
                {\color{gray}; Service is down}
                (must-cancel-booking)))
\end{Verbatim}

The make-up of the web action is very similar to a dialogue action, and the difference resides mainly in the execution. Of particular interest, the JSON payload sent to the web server is \textit{predefined to be the environment context corresponding to the action's precondition}. In other words, any aspect of the current context (e.g., the user's destination) must be included as a precondition if it is to be used as part of the invocation of the web action. This restriction greatly reduces the potential for modeling errors, and is a pattern discussed further in Section \ref{sec:exec}.

The determination of a web action corresponds to analysis of the server response. In particular, we assume that the endpoint called returns a particular format that indicates which outcome has occurred, and arbitrary web services can be automatically wrapped to mirror this behaviour.

This structure provides a powerful abstraction for dialogue designers and developers to work in tandem. Virtually \textit{any} decision support system can be embedded as a web service, and outcome determination can be as simple or as complex as the developers wish to make it. As a prime example, we have seen complex domain-specific NLU deployed as a separate web service, and a cascade of dialogue-to-web actions used as a means to process user input in a more sophisticated way.

\subsubsection*{System Action}
The final action type is a \textit{system action}. Unlike the previous two, the system actions do not have a grounded notion of execution. Rather, they can be seen as simple logic-based actions internal to the agent's own reasoning. The specification of the action in the model acquisition interface provides a rich syntax of arbitrary \textit{conditions} for the outcomes (one per), and the determination of the action occurs in a very specific manner: the list of outcomes is processed in order, 
and the first outcome with a satisfied condition is considered to be the one that occurs.

In other words, the system actions allow the designer to write declarative bookkeeping components that represent an extended if-then-else construct. 
During user testing, we found that system actions are frequently used with just one or two outcomes: either to propagate information or do a simple check and set flags as a result. An example PDDL snippet of this type of action is as follows:

\begin{Verbatim}[frame=single,framesep=5mm,commandchars=\\\{\}]
{\color{gray}; Logic action to set high/low flags}
(:{\color{blue}action} assess-temperature

    :{\color{blue}precondition} (\textbf{and} (have-temperature))

    :{\color{blue}effect} (\textbf{oneof}
                {\color{gray}; temperature > 100}
                (\textbf{and} (temp-high) (\textbf{not} (temp-low)))
                
                {\color{gray}; temperature <= 100}
                (\textbf{and} (temp-low) (\textbf{not} (temp-high)))))
\end{Verbatim}

Note the non-standard interpretation of the outcomes for this final action type: in reality, more than one condition may hold, but \textit{only the first} will be the one adopted during execution. In essence, we have allowed for a limited form of imperative style specification in a declarative framework. We found that certain key imperative notions were essential to include in order to ease the difficulty of specifying everything in a purely declarative way. This insight can also be seen below in the \textit{forced followup} enhancement we make to the model.

The specification of the individual conditions are provided in the model acquisition interface, and evaluated using our custom executor. As such, the actual conditions need not (and indeed should not) appear in the abstract PDDL model of the action's effect.

\subsection{Modeling Enhancements}
\label{subsec:sugar}

With the core encoding in Section \ref{sec:core-encoding}, we are able to encode the full range of dialogue agents we set out to cover. That being said, there are a variety of common patterns in dialogue design that warrant a custom interface and planning abstraction. We cover four of the most important ones here.

\subsubsection*{Slot Filling}
Perhaps the most frequently used pattern of dialogue design (at least under the umbrella of declarative specification) is \textit{slot filling} \cite{DBLP:conf/interspeech/GoddeauMPSB96}. The general idea is that a particular variable can be assigned by asking the user about its value. Rather than require the dialogue designer to write custom actions for every variable that can be slot filled, we allow them to simply check a box confirming that this should be possible along with the utterance that should be used to query the user. If enabled, an action similar to the following is created:

\begin{Verbatim}[frame=single,framesep=5mm,commandchars=\\\{\}]
{\color{gray}; Automatically generated slotfill action}
(:{\color{blue}action} slotfill-name

    :{\color{blue}precondition} (\textbf{not} (have-name))
    
    :{\color{blue}effect} (\textbf{oneof}
                {\color{gray}; Extracted information ok}
                (have-name)
                
                {\color{gray}; Failed, and will retry}
                (\textbf{not} (have-name))))
\end{Verbatim}

\subsubsection*{Contextual Entity Extraction}
The execution engine we use to determine user responses to dialogue actions is Watson Assistant (WA). Because of our choice to use WA, we are able to take advantage of many of the features that the product offers; a prime example being the range of system entities that the variables may correspond to (cf. Section \ref{sec:background-nl}). Another example is the sophisticated natural language processing known as \textit{contextual entity extraction} (CEE).

With CEE, we can specify a short candidate list of example utterances, and have WA generalize to all combinations of the mentioned entities. As an example, a dialogue action that asks the end user ``Can you tell me about your trip?'' can be specified using a list of utterances such as:

\begin{itemize}
\vspace{-5pt}
\item \textit{I will travel from \$src to \$dst on \$dates}
\vspace{-5pt}
\item \textit{I will fly to \$dst.}
\vspace{-5pt}
\item \textit{I'm traveling from \$src to \$dst}
\vspace{-5pt}
\item \textit{I want to take a trip on \$dates}
\end{itemize}

Note that while there are $2^k$ possible responses that use some subset of the $k$ variables, we only need to specify a short list of examples so that the natural language processing can identify the surrounding context for extracting an entity. A response such as ``\textit{I will fly to Toronto on May 15th}'', although not provided in the list above, is successfully handled by the system.

To take advantage of this advanced feature in Watson Assistant, the model acquisition interface allows for this compact specification of example utterances and we subsequently generate PDDL such as the following automatically:

\begin{Verbatim}[frame=single,framesep=5mm,commandchars=\\\{\}]
{\color{gray}; Open ended question about trip details}
(:{\color{blue}action} cee-extraction

    :{\color{blue}precondition} (\textbf{and} (\textbf{not} (have-src))
                       (\textbf{not} (have-dst))
                       (\textbf{not} (have-dates)))
    
    :{\color{blue}effect} (\textbf{and} (\textbf{oneof} (have-src) (\textbf{not} (have-src)))
                 (\textbf{oneof} (have-dst) (\textbf{not} (have-dst)))
                 (\textbf{oneof} (have-dates) (\textbf{not} (have-dates)))))
\end{Verbatim}

This feature is a prime example of how key aspects of syntactic sugar for the user of the system can greatly impact the quality of the experience, despite it not changing the expressivity of the underlying planning language.

\subsubsection*{Forced Followups}
While we strongly advocate for viewing complex agents in a declarative setting, there are some limited forms of imperative design that are hard to escape. One such example is a \textit{forced followup}. On a particular outcome of an action, if we want to be sure that the very next action is pre-defined, then doing so manually with the encoding in the declarative setting is extremely cumbersome. This is because all other actions must be ``forbidden'' in some way.

The model acquisition interface allows for a very straightforward implementation of this: each outcome can optionally specify what action is forced to follow. The resulting encoding demonstrated on the \textsc{CheckAvailability} action demonstrates the new precondition (action must be enabled) and outcome alterations (by default all actions are enabled, and outcomes with a forced followup leave all but one action disabled):

\begin{Verbatim}[frame=single,framesep=5mm,commandchars=\\\{\}]
{\color{gray}; Open ended question about trip details}
(:{\color{blue}action} check-availability

    :{\color{blue}precondition} (\textbf{and} (have-destination)
                       (have-travel-dates)
                       {\color{gray}; Every action requires for it}
                       {\color{gray}; to be enabled to execute}
                       (can-do_check-availability))
    :{\color{blue}effect} (\textbf{oneof}
    
                {\color{gray}; Dates work fine}
                (\textbf{and} (travel-ok)
                    {\color{gray}; Enable all actions by default}
                    (can-do_inform-service-down)
                    (can-do_ask-destination)
                    (can-do_slotfill-name)
                    ...)
                    
                ...
                
                {\color{gray}; Service is down}
                (\textbf{and} (must-cancel-booking)
                
                     {\color{gray}; Enable the next action}
                     (can-do_inform-service-down)
                     
                     {\color{gray}; Disable all others}
                     (\textbf{not} (can-do_ask-destination))
                     (\textbf{not} (can-do_slotfill-name))
                     ...
                )))
\end{Verbatim}

While this seemingly increases the complexity of the PDDL model, note that (1) these details are hidden entirely from the dialogue designer; and (2) the planning systems we employ are not affected by the increased modeling complexity at all. This is a prime example of how more complex modeling techniques can be used to good effect for surfacing powerful features to the designer.

As an extra variant to this enhancement, we also allow for the first action of any plan to be specified (accomplished by disabling all actions except the one we wish to begin with). This is useful to ensure that a standard greeting is provided in the conversation.

\subsubsection*{Partial Observability as 3-Valued Logic}
The final model enhancement is to extend the boolean interpretation of variable assignment (i.e., do you know the value or not) to a 3-valued logic (adding the value of ``maybe knowing'' a variable). This functionality is useful if we would like to allow for dialogue that \textit{confirms} certain variables rather than solicit them from scratch via slot filling.

As an example, assume a user's profile indicates that they are based in Boston. When they strike up a conversation with a dialogue agent to book travel, instead of assuming we don't know their source location, we can assign the context variable of their source to be ``Boston'', and the planning abstraction assumes that this is ``maybe known'' (i.e., the fluent \verb|(maybe_have_src_location)| holds). A dialogue action corresponding to a compiled slot fill is only allowed when both of the fluents \verb|(maybe_have_src_location)| and \verb|(have_src_location)| are false. On the other hand, a \textit{confirm} action for the source location is allowed when the former fluent is true, and sends the user a message such as ``Will you be traveling from Boston?''.

Aside from initial configuration of the 3-valued logic, we have found it quite useful to have the third option in other aspects of dialogue as well. One example is when the natural language understanding is only partially confident in the entity it has extracted, in which case we can allow the agent to confirm the value. Another example is in agents that do similar tasks repeatedly, and information could potentially be carried over: instead of soliciting the information repeatedly, the agent confirms if the information is still valid.

Beyond offering a more personable and fluid conversation, this feature also greatly simplified the task of natural language understanding as well: it is far easier to detect an affirmative or negative response than to attempt entity extraction.

\subsection{Discussion}
We have only scratched the surface as to what is possible with the planning view on dialogue design. 
A more complete example that touches on many of the aspects in this section can be found at:

%  Read-only: http://editor.planning.domains/#read_session=cH12680BRp
% Read/Write: http://editor.planning.domains/#edit_session=uVIwj4V8xKeEGvd

\begin{center}
\url{http://editor.planning.domains/#read_session=cH12680BRp}
\end{center}

One common thread we have discovered is the need for \textit{dialogue-specific abstractions}. All but the third enhancement mentioned in the previous subsection is a prime example of this. While a domain-independent planner is used as a computational core (and rightfully so, as the domain specification will change for every dialogue agent), the common encoding patterns are abstracted away and presented to the dialogue designer as compact methods to accomplish common tasks. This hints at a broader need from the planning community to investigate encoding patterns for general classes of a domain setting.

\subsubsection*{Fairness}
Non-deterministic planning typically involves an assumption of \textit{fairness}: loosely defined as every outcome of a non-deterministic action having some chance of occurring. In our setting, this can manifest in a number of modeling fallacies. For example, asking the end user the same question multiple times will not achieve the desired outcome in the limit. Nor will applying a system action or a web action again and again.

While we do employ a planner that assumes fairness, we leave it to the dialogue designer to make use of standard encoding methods for avoiding the incorrect modeling of unfair actions; e.g., by having a guard on an action that is removed as a result of applying it (thus making the action only applicable once until the guard is reset). There are situations where fairness is not only safe to assume, but desired as well. One example is an action that polls a web service which is busy and proceeds once it becomes available.

Moving forward, we may build on techniques that blend assumptions of fairness and unfairness on the actions \cite{cam-mci-knowpros16}. A common design pattern we could surface to the dialogue designer is to (optionally) allow for actions to be executed a bounded number of times, at which point another course of action must be taken. Similar to the forced followup feature, this can manifest in a complicated planning model, but the exposed interface to the dialogue designer remains simplified 
in the form of a single option to check.

\subsubsection*{Probabilities}
In this paper, we only discuss a qualitative notion of uncertainty. However, there is an obvious extension to a probabilistic setting that can be used. If the individual outcomes have a probability associated with them, then plans of maximum expected reward become feasible with a different planner at the computational core.

This is a direction out of scope for this paper, and currently under active exploration. One benefit worth pointing out, however, is that the probabilities associated with each outcome can play an important role in refinement of the original model. In the context of Figure \ref{fig:overview}, analytics from end-users can be aggregated and reflected in the likelihood of action outcome probabilities. This in turn can be used to generate a new plan with a higher chance of successful outcomes.

While it changes the solver required, and opens the door to interesting objective functions and model refinement opportunities, the modeling and execution process remains nearly identical to a setup where there are no probabilities associated with the actions. For clarity, we will focus entirely on the qualitative case for this work.

%% file: sections/mai.tex
\section{Model Acquisition Interface -- \mai}
\label{sec:mai}

Going back to the original motivation of declarative
design of goal directed dialogue agents (for
tasks such as customer support), 
we conceive of the following desiderata for the 
modeling interface:

\begin{itemize}
\item[-]
The designer must be in control of the behavior of the bot towards
the end-users once it is deployed. This means that the specification of the bot
must be interpretable to the designer, and also readily debuggable and editable 
as desired by the designer.\footnote{This more or less precludes the direct use of state-of-the-art deep learning based end-to-end dialogue systems as the substrate for the bot specification. However, we will see later in Section~\ref{sec:future} how our approach provides pathways towards effective integration of data-driven approaches along the course of the design process.}
\item[-]
The designer should be able to connect the bot's actions to system functions or calls to API endpoints (e.g. to issue a ticket or place an order).
\item[-]
Finally, and perhaps most importantly, in order to be able to support
complex interaction patterns between the end-user and the bot,
the bot designer must be able to completely describe the bot's capabilities
without having to manually specify the entire dialog tree.
\end{itemize}

We will now describe in detail how
our model acquisition interface -- \mai\ (Declarative Design of Dialogue Agents
for Watson Assistant) -- goes about making these
capabilities available to the bot designer through the paradigm of
declarative modeling.
% An overview of the project can be viewed at {\protect\url{ibm.biz/mai-video}}.
A snapshot of the \mai\ interface (annotated with its salient features) 
is shown in Figure \ref{fig:d3wa-inspection}.
This concerns the left-most step of our overall pipeline illustrated
in Figure \ref{fig:overview}.
In the following discussion, we will provide a quick tour of the core features
of the interface and outline the journey of the bot designer from start
to eventual deployment of the bot.
We will touch briefly on the {\em beta} features later in Section \ref{sec:future}.

\subsection{The Anatomy of a Declarative Specification on \mai}

% As we mentioned before, the state of the art for the design of dialogue agents
% involves manual specification of the entire dialog tree. 
% This process is imperative, in having to enumerate all sequences of 
% possible bot behavior. 
The start of the declarative modeling process starts with the
bot designer putting themselves in the shoes of the agent
and conceiving of {\bf variables} that the agent needs to keep track of 
and {\bf actions} that the agent is capable of executing.

\begin{itemize}
\item[-]
{\bf Variables:}
The panel on the left in Figure \ref{fig:d3wa-inspection} shows a list of variables that the bot keeps track of during the course of an interaction.
These variables may be {\tt Booleans}, arbitrary {\tt JSON}s, or
variable types defined as {\em system entities} in Watson Assistant\footnote{System Entities in Watson Assistant: \url{http://ibm.biz/sys-entities}} that may be extracted from user utterances.
Examples of variables include credit card details (as a {\tt JSON} dictionary),
{\tt Boolean} flags determining whether certain pieces of information have been acquired, etc.
\item[-]
{\bf Actions: }
Actions appear in the central panel of \mai\ (cards in the middle of Figure \ref{fig:d3wa-inspection}) -- they define capabilities of the bot. 
They operate on the variables towards achievement of its goals.
Each action is defined from the perspective of the bot in terms of 
the information it requires to perform the action and the possible outcomes 
of performing that action.
\end{itemize}

\begin{figure}
\centering
\includegraphics[width=\textwidth]{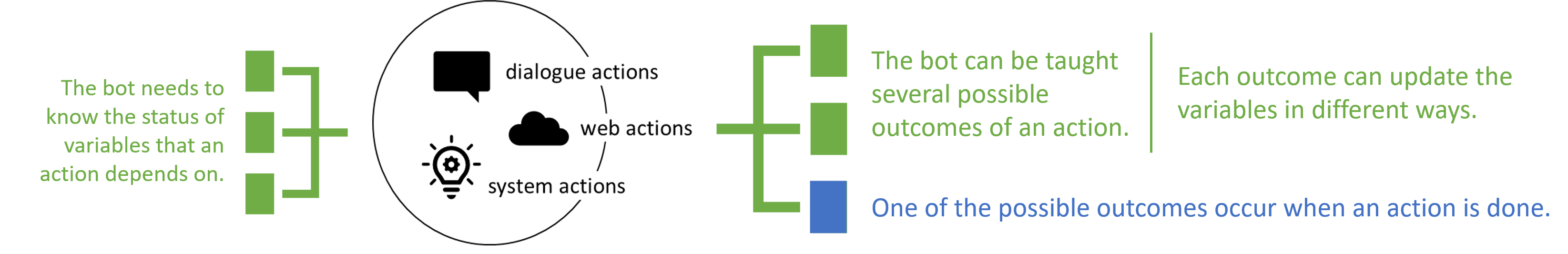}
\caption{
Anatomy of the declarative specification of an action on \mai.
This structure -- detailed in Section \ref{subsec:structure} -- 
is shared by all three actions types -- i.e. dialogue, web, and system
actions. Their subtle differences are discussed in Section \ref{subsec:types}.
}
\label{fig:dec-action-30k}
\end{figure}

\subsubsection{Action Structure}
\label{subsec:structure}

An action on \mai\ consists of the following three components (c.f. Figures \ref{fig:d3wa-inspection} and \ref{fig:dec-action-30k}):

\begin{itemize}
\item[-] {\bf Needs}
are values -- {\tt true} or {\tt false} -- or status --  {\tt known}, {\tt unknown} or {\tt uncertain} -- of variables that need to hold for the bot to be able to perform that action. For example, the credit card details would need to be known before the bot can place an order. These correspond to the 
preconditions of the action in the PDDL specification compiled in the backend.
\item[-] {\bf Outcomes} 
are the set of possible events that can occur in response to the bot executing the action. For example, if the bot calls an API endpoint to place an order, it may either receive some data back indicating success or it may receive a failure message -- this can be modeled in terms of two different outcomes, each
of which updates the world in different ways. 
The designer gets to specify as many possible outcomes as they wish --
in the PDDL specification being compiled in the background, 
these become the mutually exclusive non-deterministic outcomes for the action.
Thus, when the bot is compiled, the planner composes the full dialogue tree considering all possible outcomes, while at the time of execution, only one of the outcomes occur when an action is performed -- the job of the {\em execution monitor} (described in Section \ref{sec:exec}) is to identify and react to the outcome that has occurred. 
\item[-] {\bf Updates}
define how the values of variables change for each outcome. 
For example, if the user responds with the credit card details, then
the values of the variable corresponding to credit card details are updated,
or the status of the order may be updated after calling an API endpoint, as 
described above. The manner of assignment of variable updates largely depends on the type of an action, as described next.
\end{itemize}

\subsubsection{Action Types}
\label{subsec:types}

The interface allows three types of actions that perform different roles
in the course of an interaction between the bot and the end user. 
All of them follow the action structure introduced previously,
in addition to extra features typical to a particular action type.
Further details on the three types is provided in Section \ref{sec:core-encoding}.

\begin{itemize}
\item {\bf Dialogue}
actions are performed directly with the user via conversation.
The designer outlines how the bot expresses itself (such as in asking for credit card details) and possible user utterances in response for each possible outcome.
\item {\bf System}
or ``logic'' actions are internal to the bot and are used to make
internal inferences about the world to maintain and update state information 
(for example, in setting credit card details to {\tt known} once the bot determines membership in loyalty program). 
The designer can specify under what (logical) conditions each outcome occurs and in what order. The ability to design these actions is unique to this design paradigm -- internal actions are not accessible in chat logs and are thus inaccessible to end-to-end learning systems.
\item {\bf Web}
actions or ``cloudfunctions'' allow the bot to respond to actionable items during an interaction (for example, placing an order with the acquired card details).
This is crucial to the development of goal-oriented dialogue agents in domains such as customer support where the agent needs to perform actual tasks beyond chitchat, for example, to pull data from a user's account, set information such as usernames, passwords, etc. issue tickets, book orders, and so on.
Of course, these API endpoints still need to be implemented by a developer -- however, \mai\ provides pathways to {\em simulate} these endpoints in situ 
at design time so that the designer can chat with the bot immediately.
\end{itemize}

\noindent For dialogue actions, information extracted from the specified user utterances (using classifiers automatically trained using Watson Assistant services) make internal updates to variables in each possible outcome.
For cloudfunction actions, such updates will come from the response of the API endpoints. These updates are auto-generated and are in addition to those explicitly defined by the designer in the outcomes of an action, as described before.

\subsubsection{Enhancements: Follow-ups, Confirmations, and Slots}

As we described in Section \ref{subsec:sugar}, 
we provide specific modeling enhancements to simplify the specification
based on some commonly occurring design patterns.
Specifically, the interface houses special pathways to add actions that are designed to (1) fill slots (i.e., query for missing information) or (2) confirm values of variables when the bot is uncertain. Additionally \mai\ allows for the designer to (3) start the conversation with a particular action; (4) end the conversation on a particular outcome; or (5) force the next action as a \textit{follow-up} to a particular outcome.

Some of these features, such as the forced follow-ups, hark back to the imperative process of bot design and are left in as syntactic sugar to further streamline the design process or sometimes to make the transition to the declarative modeling paradigm easier. The designer is recommended to use them sparingly, wherever absolutely necessary.

\subsubsection{The \mai\ Python3 API}

In addition to the graphical interface described so far, we also
provide a \mai\ API in Python3 that may be used to generate a 
specification consumable by \mai\ using code. 
This provides an alternative way to interface with the infrastructure.
Crucially, the API captures the full range of functionality in MAI (rather
than simply cover the generated PDDL abstraction). The implication of this
is that all of the enhancements described in Section \ref{subsec:sugar} can
be readily configured through the API programmatically.
The API is particularly helpful in two cases:

\begin{itemize}
\item[-] 
Often we provide pathways for other abstractions for specification that
can be translated into the core \mai\ format. We do this, for example,
in the instances of \cdd\ and \dba\ ({\em beta} features) described briefly 
in Section \ref{sec:future}.
The compilations in these scenarios are done by making calls to the 
API in the backend.
\item[-] 
The API is especially useful when generating bots from 
large specifications with repeating conversation patterns
that the designer need not enumerate on the interface.
The Credit Card Recommendation domain introduced later in
Section \ref{subsec:scaleup} uses this -- there, given the 
features of a particular credit card, the specific 
details of the conversation around them can be generated
through the API. The designer does not need to repeat
these patterns for every card.
\end{itemize}

\subsection{\mai\ Build-Debug-Deploy Cycle}

Once the design is complete, the designer can then proceed to build their bot, which enables the following three functionalities:

\begin{itemize}
\item {\bf Visualize the Generated Dialogue Tree} 
Once a solution is generated from the planner, 
the designer can proceed to visually inspect the resulting dialogue tree.
This helps them to better understand the dependencies in the current
specification and identify unmodeled constraints or 
possible undesired dialogue patterns.

\item {\bf Chat with the Bot} 
The designer can also chat with the bot directly
(panel on the right in Figure \ref{fig:d3wa-inspection}) and
trace the progress of the dialogue along the visualized tree.\footnote{The latter functionality is currently in a separate interface, 
as described later in Section \ref{sec:hovor-frontend}.}
This provides a powerful debugging tool towards further
refinement of the bot specification.

\item {\bf Deploy to Watson Assistant} 
Finally, the designer can deploy the bot directly to Watson Assistant\footnote{https://www.ibm.com/cloud/watson-assistant/} --
this compilation is a good way to appreciate the massive scale-ups 
achieved in the complexity of the bot with respect to the
complexity of the specification.
\end{itemize}

\subsection{\mai\ Backend}

The \mai\ back-end compiles the specification in the GUI into two forms: (1) an abstraction of the declarative planning process that a blackbox planner can solve, producing the entire dialogue tree; and (2) the configuration that is required for deployment of the dialogue agent, including elements such as the example utterances to detect responses from the user or from calls to endpoints.

As we mentioned before, the non-deterministic planner plans with all possible
outcomes of an action while generating the complete dialog tree.
However, during execution, only one of these outcomes occurs. 
The realization of an action in implementation is done automatically from
the specification in the form of {\em determiners} which determine 
which outcome has occurred. 
For example, for dialogue actions, the determiners rely heavily on natural language processing services from Watson Assistant.
The \hovor\ plan executor orchestrates this entire process 
and situates the bot in the right location in the dialogue tree. 
We describe the design of the executor in the next section.

%% file: sections/executor.tex
\begin{figure}[t]
    \centering
    \includegraphics[width=0.8\linewidth]{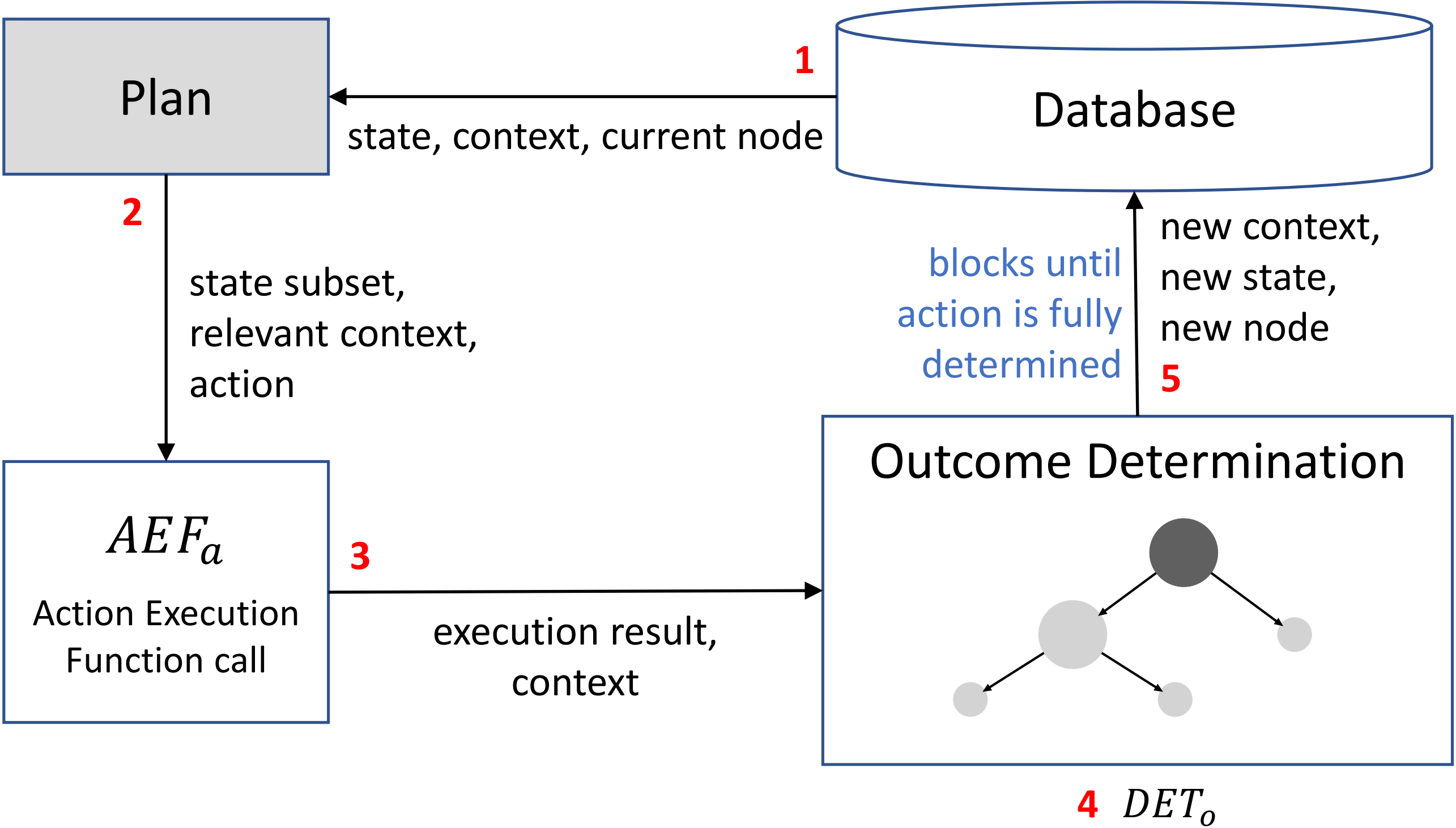}
    \caption{Architecture for executing contingent plans.}
    \label{fig:architecture}
\end{figure}

\section{Plan Executor}
\label{sec:exec}

Our aim here is to provide a coherent and effective strategy for deploying dialogue agents that are based on the execution of a contingent plan. In this section, we identify some of the key challenges that arise in building an executor of such plans and propose a solution to each of them.
The high-level architecture we propose is in Figure \ref{fig:architecture}. The execution involves the following phases:

\begin{enumerate}
    \item At every iteration of the execution monitor, we have the state of the world (relevant to the planner), a context of variable assignments (used by the executor, not needed by the planner), and the plan's current node $n$ along with the corresponding action $a = \actmap{n}$.
    \item The executor retrieves the relevant context and state for action $a$ (Section \ref{ssec:exec-vs-det}).
    \item The action execution function ($\exec{a}$) is called with the filtered context.
    \item The action's effect is determined and realization chosen (Section \ref{ssec:complex_outcome_determination}).
    \item The state and the context are updated (Section \ref{ssec:complex_outcome_determination}).
\end{enumerate}

% General Framework
% \begin{figure}[t]
%     \centering
%     \includegraphics[width=\linewidth]{figs/execution_monitor_cropped5.pdf}
%     \caption{Architecture for executing contingent plans.}
%     \label{fig:architecture}
% \end{figure}

To motivate the type of complex actions we wish to execute, consider an example scenario where a the trip booking agent is going
to finally put in a reservation for the hotel. 
Figure \ref{fig:pddl-example} shows an example PDDL representation of an action which puts in the request for the hotel reservation, and Figure \ref{fig:effect-example} shows the accompanying effect structure. Notice that some of the non-determinism is independent (e.g., the accessibility of the account and whether the request gets confirmed immediately), and some of the non-determinism contains a dependency (e.g., whether the card works or not is only known if the account is accessible). We expand on this further in the following sections.

\begin{figure}[t]
\centering
\begin{Verbatim}[frame=single,framesep=5mm,commandchars=\\\{\},fontsize=\small]
(\textcolor{blue}{:action} book_hotel
  \textcolor{gray}{; can access account and know card details,}
  \textcolor{gray}{; not attempted booking hotel yet}
  \textcolor{blue}{:precondition} (\textbf{and} (account_accessible)
                     (\textbf{not} (card_unknown))
                     (\textbf{not} (attempted_hotel_booking)))
  \textcolor{blue}{:effect} (\textbf{and}
            \textcolor{gray}{; attempted hotel booking}
            (attempted_hotel_booking)
            (\textbf{oneof}
              \textcolor{gray}{; booking is confirmed}
              (hotel_booking_confirmed)
              (\textbf{and}
                \textcolor{gray}{; confirmation is pending}
                (\textbf{not} (hotel_booking_confirmed))
                (hotel_booking_pending)))
            (\textbf{oneof}
              \textcolor{gray}{; account no longer accessible}
              (\textbf{not} (account_accessible))
              (\textbf{and}
                \textcolor{gray}{; account is still accessible}
                (account_accessible)
                (\textbf{oneof}
                  \textcolor{gray}{; but card does not work anymore}
                  (\textbf{not} (card_unknown))
                  \textcolor{gray}{; card information works}
                  (card_unknown))))))
\end{Verbatim}
\caption{Part of a PDDL specification of an action for putting in a hotel reservation at the end of the trip booking conversation, which demonstrates the various complexities of nested non-deterministic effects in the context of modeling web actions.}
\label{fig:pddl-example}
\end{figure}

\begin{figure}[t]
    \centering
    \includegraphics[width=\linewidth]{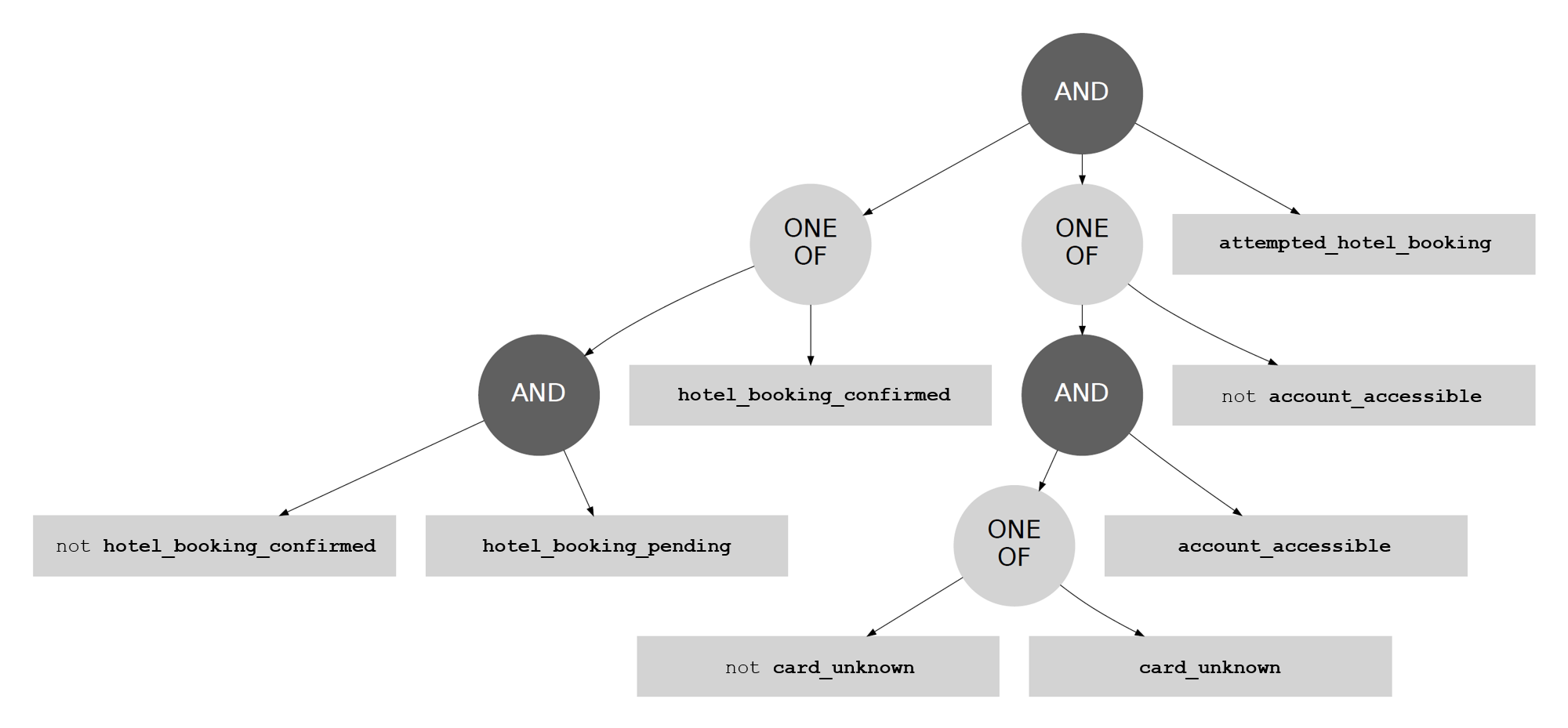}
    \caption{Effect tree of the action in Figure \ref{fig:pddl-example}.}
    \label{fig:effect-example}
\end{figure}

As alluded to in the previously, we can view the effect of an action as an \textit{and-or} tree (illustrated for our example in Figure \ref{fig:effect-example}). While this analogy is useful for some aspects, such as defining the realizations of an effect as $\realizations{}$, it is important to recognize the distinction from a Boolean formula represented as an \textit{and-or} tree: \textbf{the effect of an action is not a Boolean function to be evaluated}. The biggest ramification of this is that we evaluate the effect in a top-down manner (starting with the root) rather than bottom-up. We elaborate on this point further in Section \ref{ssec:complex_outcome_determination}. Also, note that unlike a common \textit{and-or} graph, we cannot simplify it by operations such as merging parent-child \textit{or-or} node relations or collapsing seemingly tautological subtrees (such as the bottom \textit{oneof} node in Figure \ref{fig:effect-example}), as each \textit{oneof} node has a unique determiner associated with it.

\subsection{State-Context Mapping}
\label{ssec:state-context-mapping}

\hide{          % ALL DONE %
\textbf{To discuss:}
\begin{itemize}
    \item Context is an assignment of values we maintain for execution, but the planner doesn't care about.
    \item State are the fluents that the planner knows and cares about.
    \item We find a reasonable abstraction so that the planner only needs to know the minimal amount of information to get by.
    \subitem * Based on the global example we can demonstrate it.
    \item The state and context should be aligned for all reachable states -- more on this when we talk about the role of determination.
    \item \textbf{(have-*)} fluents mapping to actual context variables. Only as an example (this paper doesn't need to go into the dialogue setting).
    \item \textbf{(maybe-have-*)} a controlled way of expressing uncertainties
\end{itemize}
}

Beyond the planner's abstract view  of the world (the \textit{state}), a real-world system usually needs to retain contextual information needed for the practical execution of the plan. This brings the first key challenge that must be overcome: we need to track the correspondence between the state and the context. The execution state is a set of fluents that hold and are important for the planner to decide which actions can be executed. The context $C$ is an aligned assignment of values $C: \fluents \rightarrow Dom$, where $Dom$ is an arbitrary (and possibly open ended) domain. Note that as we have defined it, there is a one-to-one mapping between the state fluents and the context variables. This is by no means the only approach one could adopt, and we discuss the difficulties with generalizing this interpretation below.

The alignment can be illustrated with our running example of the 
\texttt{book\_hotel} action in the trip planning conversation: the state contains a fluent $\texttt{(hotel\_booking\_pending)} \in \fluents$ denoting whether the booking has been attempted but not
confirmed yet. From the planner's viewpoint, the only important result from attempting a hotel booking is whether the system knows the status of the reservation (regardless of what stage it is in). However, the booking status itself may play a role in the execution and determination of other actions, so the context value associated with the fluent (e.g., $C(\texttt{(hotel\_booking\_pending)}) = \textit{Checking Availability}$) may become important. 
For example, there may be another action, \texttt{check\_booking\_status} that has a precondition \texttt{(hotel\_booking\_pending)} and non-deterministically results in having either \texttt{(checking\_availability)} or \texttt{(not (checking\_for\_automatic\_upgrades))} (the former could act as a precondition for returning to the user for 
a confirmation for a free upgrade).

%
%Skipping for now, since our "have" example doesn't make sense with the "maybe" notion.
%
%When more fine-grained information about context values for planning is required, new fluents need to be added. E.g. if planner needs to select its actions based on confidence levels, ${maybe\_have\_location(src)}$ can be added and used for deciding whether the source location should be confirmed or not.

The advantage of splitting the execution information is twofold. 
First, it situates the planner at the right level of abstraction with the context maintained separately from the state.
Second, it allows an interconnection between the plan and complex objects (like a web call result, agent coordinates, etc.) which would be impractical to represent in the planning domain.
Maintaining the context alignment for all reachable states is one of the challenges in outcome determination, and is described further in Section~\ref{ssec:complex_outcome_determination}.

\subsection{Action Execution -vs- Determination}
\label{ssec:exec-vs-det}

\hide{          % ALL DONE %
\textbf{To discuss:}
\begin{itemize}
    \item General philosophy: split an action's execution into two phases; (1) is the execution of the action itself (system call, whatever), and (2) is the realization of the result.
    \item Phase (1) is the initiation of the action, and actually invokes whatever real-world process is happening.

    \item Both execution and determination can only view (TODO: updates should be also somewhat limited) the subset of the context / state that appears in the precondition: a guiding principle to avoid executor errors (model should reflect assumptions on execution / determination).
    \item TODO: I'm not sure whether context and state updates belong here or to the following subsection (assuming the later).
\end{itemize}
}

The aim of contingent plan execution is running action callbacks associated with each state to achieve a specified goal state.
However, in a non-deterministic world, an action can cause one of multiple effects in the state and context. 
%Therefore, an essential part of the plan execution is deciding which of the effects has happened, i.e. the outcome determination.
Moreover, in the real world, the effects may not be fully observable and may be too complex to be modeled perfectly.
Therefore, an essential part of the plan execution is deciding which of the effects best describes the real world change. This is the process of \emph{outcome determination}.

This leads us to a natural decomposition of action execution into two phases (Steps 3 and 4 in Figure \ref{fig:architecture}). First, running the function callback that implements the real process in the outside world (e.g. calling a web service), which gives the system a function call result (e.g. a response code with a payload of information from a web service).
Second, the outcome determination, which processes the function call result to update the execution state and context.
Outcome determination is a complex multistep process described in detail in the following section.
% Section~\ref{ssec:complex_outcome_determination}.
At a high level, it involves running the appropriate subset of callback functions in the appropriate order to establish what single realization should be used to update the state of the world and the solution status.

The separation between action execution and outcome determination is not just conceptually attractive from the standpoint of clearly separating distinct functionality in the implementation; it also provides a natural means of encouraging better declarative models. With the evaluation of an effect focused on computing what has changed (as opposed to making additional changes to the real world), complex and error-prone action models are avoided in lieu of multiple actions, each with a clearly defined purpose. The core anti-pattern that is avoided is the strategy of embedding aspects of an agent policy directly within an action effect (e.g., if a determiner selects one branch, then further actuation occurs that influences the environment).

In theory, the callback function and outcome determiners could be given the entire state and context. 
However, such a practice would result in an error-prone and hard to debug system.
Therefore, \textit{only the context subset claimed in an action's precondition is accessible for the action execution}: $C_a = \{C(f) | f \in \pre a \}$. This drastically reduces the potential for model mismatch due to modeling errors between the planning view of state, and full view of context. Providing the full context and state would be akin to using global variables exclusively in software development -- a practice largely viewed to be error-prone and undesirable.

\subsection{Complex Outcome Determination}
\label{ssec:complex_outcome_determination}

\hide{          % ALL DONE %
\textbf{To discuss:}
\begin{itemize}
    \item Determination needs to find realization based on action effect specification (TODO reference to the PDDL example)
    \item Determination is a non-trivial task $\rightarrow$ decompose it to and-or graphs 
    \item Data revision (i.e. context updates) for the outcome determination
    \item Effective execution (parallelism) can be used thanks to the decomposed determination
    \item Outcome ranking as an approach to cope with Partial observability
    \subitem * Full action determinations can be driven by recursive rankings (with performance impact however) or with a limited look-ahead
    \subitem * "Outcome determination learning" can be defined with the rankings (TODO how deep want we go into it?)
    \item Visual example of the graph that mirrors a complex effect written in PDDL.
    \item Determination need not (and at times cannot) be done exhaustively -- instead you work top-down and determine only what needs to be.
    \item The state/context do not change until phase (2). This is why we need the assumption of non-overlapping or complex durative actions.
    \item Determination can be a long-running process in itself -- but the separation of execution (i.e., ``do something'') and realization (i.e., ``what's changed'') should be maintained. That is, in order to avoid escalating complexity of individual action specifications, and the potential for unforeseen side-effects, an action should correspond to at most one physical transition in the real-world setting.
\end{itemize}
}

The outcome determination is a fundamental part of action execution within a contingent plan. 
The role of determination is to decide which realization $r \in \realizations{a}$ of the action $a$ has occurred after the action callback $\exec{a}$.
The action realization $r$ is made up of $\add{r}$ and $\del{r}$ sets.
During the outcome determination, those sets are recursively calculated from the action effect tree, and use the $\determiner{}$ callbacks to compute these values for updating both the execution state and context.

The actual result of running $\determiner{}$ is for one of the outcomes it represents to be selected. To get a grounded sense of what occurs in these implementations consider the three example actions given in Section \ref{sec:core-encoding}:

\begin{enumerate}
    \item \textsc{AskDestination}: The determiner for this dialogue action's effect will use the utterance of the end-user to detect which intent (e.g., having a valid destination provided) actually fired. This includes the fallback option which is represented as its own outcome.
    \item \textsc{CheckAvailability}: The determiner for this web action's effect will assess the response from the hotel booking web service. Depending on the server response, the appropriate outcome will be chosen.
    \item \textsc{AssessTemperature}: As a logic-based system action, the determiner in this case will go through the list of conditions one after another, and stop on the first that matches. Here, the first condition will simply compare the context variable for temperature with the value of 100.
\end{enumerate}

Note that each of the action types have their own type of determination: to simplify exposition of our work, the actions only list a single determiner with a specific purpose. However, the executor we have constructed is fully general and allows for a heterogeneous mix of determiner types (pre-existing and newly defined).

\subsubsection{Recursive Effect Processing}
The effect tree consists of nodes with two kinds of operators.
First, the \textit{and} operator $\effand {\varphi}$ with sub-formulae ${\varphi_i}$ can be recursively calculated as:
\begin{gather*}
    \add{\effand {\varphi}}  = \bigcup_{\varphi_i} \add{\varphi_i}  \hspace{2em}
    \del{\effand {\varphi}}  = \bigcup_{\varphi_i} \del{\varphi_i}
\end{gather*}
Second, the \textit{or} operator $ \effoneof {\varphi} $ and its selected child $\varphi_j$   :
\begin{gather*}
    \add{\effoneof {\varphi} }  = \add{\varphi_j}
    \hspace{2em}
    \del{\effoneof {\varphi}}  = \del{\varphi_j}
\end{gather*} 
The effect tree leaves $\varphi_L$ directly define the $\add{\varphi_L}$ and $\del{\varphi_L}$ sets.
With this recursion defined, the action realization update sets are calculated as update sets for the effect's top-level root node.
The crucial part of outcome determination is the $ \effoneof {\varphi_j} $ selection. 
In practical systems, the selection can be a time consuming service call. 
For instance, the agent can use an outcome determiner that performs a remote call to a tailored NLU processor (e.g., one that detects sentiment of the utterance).
%
%deployed recognition model for detecting entities in a scene (e.g. one detecting whether there is an obstacle in the driveway or one detecting whether it is already dark outside to require headlights).
%
%The model recognizes intents and entities in user input (obtained during the previous action \textit{listen\_to\_command}) that the agent needs to decide how the state should be changed.
%
Such calls can take a long time to execute (due to network latency and complexity of the computation), and therefore exhaustive calculation over the whole tree (which can be full of such expensive calls) might not be suitable.

Thanks to the tree structure, effects can be processed top-down, evaluating only the nodes that can contribute to the realization. 
For example, each \textit{oneof} node needs only one sub-tree to be fully evaluated, and processing of that one sub-tree will only begin once the determiner for the \textit{oneof} node has identified that it is the right one to proceed with.
On the other hand, all of the sub-trees of an \textit{and} node must be evaluated, which can be done in parallel (recall that the resulting update sets are not contradicting by definition). This process can be captured by the following algorithm:

\makeatletter
\xpatchcmd{\algorithmic}{\itemsep\z@}{\itemsep=0.5ex plus2pt}{}{}
\makeatother

\begin{algorithm}
\caption{Parallel Nested Determination algorithm}\label{alg:parnes}
 \footnotesize
\begin{algorithmic}[1]
\Procedure{\textsc{ParallelNested}}{$node$}
\If{$node.type=leaf$}
    \State \textsc{ProcessLeaf}$(node)$
\ElsIf{$node.type=oneof$}
    \State $child = DET_{node}()$\Comment{Run determiner}
    \State \textsc{ParallelNested}$(child)$
\ElsIf{$node.type=and$}
    \State apply\_async(\textsc{ParallelNested}, node.children)
\EndIf
\EndProcedure
\end{algorithmic}
\end{algorithm}

Note that on line 8, the children of an \textit{and} node are concurrently processed, while on lines 5-6, the determiner for a \textit{oneof} node is run until completion before recursing. Line 3 encapsulates the recursive $\add{}$ and $\del{}$ computation.

Retaining the full complexity of action effects thus gives us two key improvements in efficiency: (1) the ability to avoid evaluating sub-trees that correspond to outcomes a determiner deems did not occur; and (2) the ability to run determiners in parallel when they represent sibling sub-trees of an \textit{and} node in the effect graph.

\subsubsection{Dependencies Between Effects}
There is another consideration which makes our recursive tree structure necessary to leverage parallelism. Some of the determiners may be able to start evaluating only once some other determiners have finished executing. For instance, in our hotel booking example, whether the card information is valid is only relevant if the account is accessible. In more extreme examples, a determiner may not be executable at all if its parent determiner did not resolve in a way that enables the child (i.e., having the right outcome selected). Such dependencies can prevent full parallel execution of all the determiners simultaneously in a flattened structure while our recursive tree structure does allow us to leverage at least partial parallelism in such cases.

\subsubsection{State and Context Updates}
With the realization update sets prepared, the new state after the action execution can be calculated as usual (assuming realization $r$): $S_{i+1} = (S_i \setminus \del{r}) \cup \add{r}$.
%\begin{gather*}
%    S_{i+1} = (S_i \setminus \del{r}) \cup \add{r}
%\end{gather*} 

Context $C$ also has to be updated during the outcome determination. 
The new context values are produced by the determination process directly along the tree traversal described above (i.e., as part of the $\determiner{}$ callback functions), and can additionally make use of the $\exec{}$ action response.
Our running example demonstrates one possible update: the status of the pending hotel booking will be set during the determination process given the information computed by $\exec{\texttt{book\_hotel}}$.
%
%Various outcome determiners are expected to provide many types of context values. Example of an outcome determiner can be an entity extractor providing a specific entity type for the context and selecting $\varphi_i$ according to the entity presence in the function call result features $z$.
%
Formally, a realization $r \in \realizations{a}$ will have
%an associated set of
context updates $C_r$ defined as:
\begin{align*}
    \set{C(f) = val \st f \in \add{r}} \cup \set{C(f) = \bot \st f \in \del{r}}
\end{align*}

The assignment of $C(f) = val$ is defined by the determiner callback for the effect tree's leaf nodes, $\determiner{\varphi_f}$ (not all fluents have context necessary for execution, in which case we set $C(f) = \bot$). Notice that such updates force the precise alignment of context and state, which is necessary for the system to function properly during execution. 

%The fluents and context are updated solely from $\exec{}$ action response that can be understood as a measurement of real world objects.
%This way, we avoid complex (circular) dependencies between fluents and context values that can be hard to manage as shown in TODO.

Dependencies introduced by real world mechanics can cause difficulties for context value updates.
For instance, fluent \texttt{(have\_location)} and its context value $[city,state]$ may represent the user's location 
(and the validity of the information in the user's account).
Then, fluents like \texttt{(have\_city)} and \texttt{(have\_state)} and their context values $[city]$ and $[state]$ correspond to separate position sub-components.
An action changing a single sub-component for fluent \texttt{(have\_city)} must also update context value of \texttt{(have\_location)} which breaks action encapsulation and may lead to errors during development.
Effective modeling of such dependencies is an interesting challenge that we defer to future work.

\subsubsection{Completing Determination}
Once we have completed the determination process for action $a$, we are left with a new state $s$, context $C$, and realization $r \in \realizations{a}$. Given the current node $n$ of the contingent plan, we compute the updated node as $n' = \edgemap{n}{r}$. The final step of the executor is to store all of the newly computed information ($s, C, n'$) in a centralized database (Step 5 in Figure \ref{fig:architecture}).

%% file: sections/hovor-frontend.tex
\section{Diagnostic Executor Frontend}
\label{sec:hovor-frontend}

\subsection{Overview}
Our framework offers an additional diagnostic frontend interface to the executor aimed at aiding dialogue designers in leveraging usage information to iteratively improve on the agent’s design. The complex web generated by dialogue states, intents and outcomes becomes harder to grasp and debug as the dialogue scope increases. Therefore, the interface displays these contingent plans generated by our planner alongside the dialogue flow between the agent and the user with interactive features to enhance the understanding of the process as it unfolded during the original conversation.

\begin{figure}[ht]
    \centering
    \includegraphics[width=\linewidth, height=8cm]{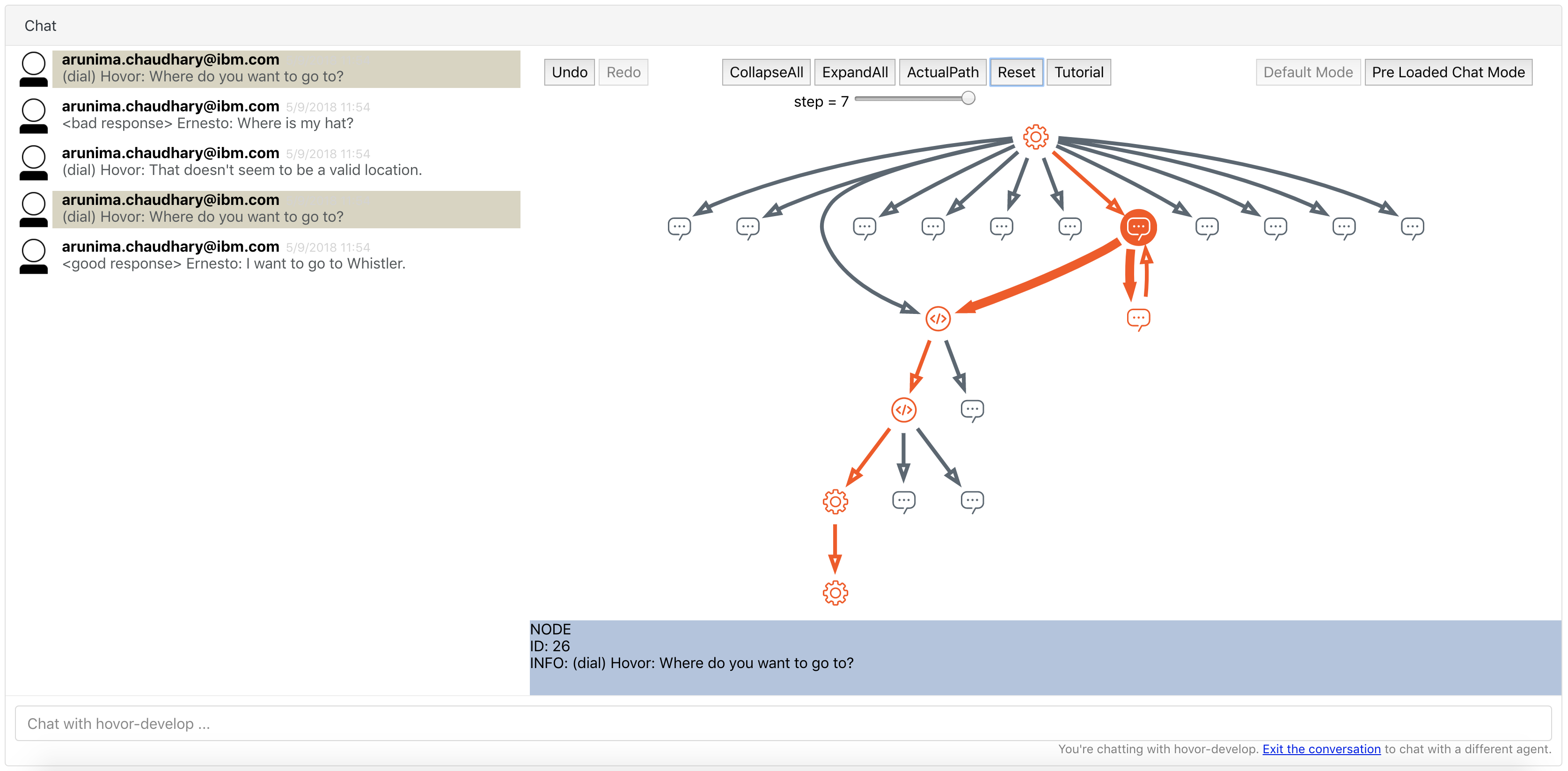}
    \caption{Screenshot of the tool with a conversation and trace highlighted. The emphasized edge in the right view matches two utterances in the left, as that edge has been traversed twice.}
    \label{fig:hovor-frontend}
\end{figure}

\begin{figure}[ht]
\centering
\includegraphics[width=0.5\columnwidth]{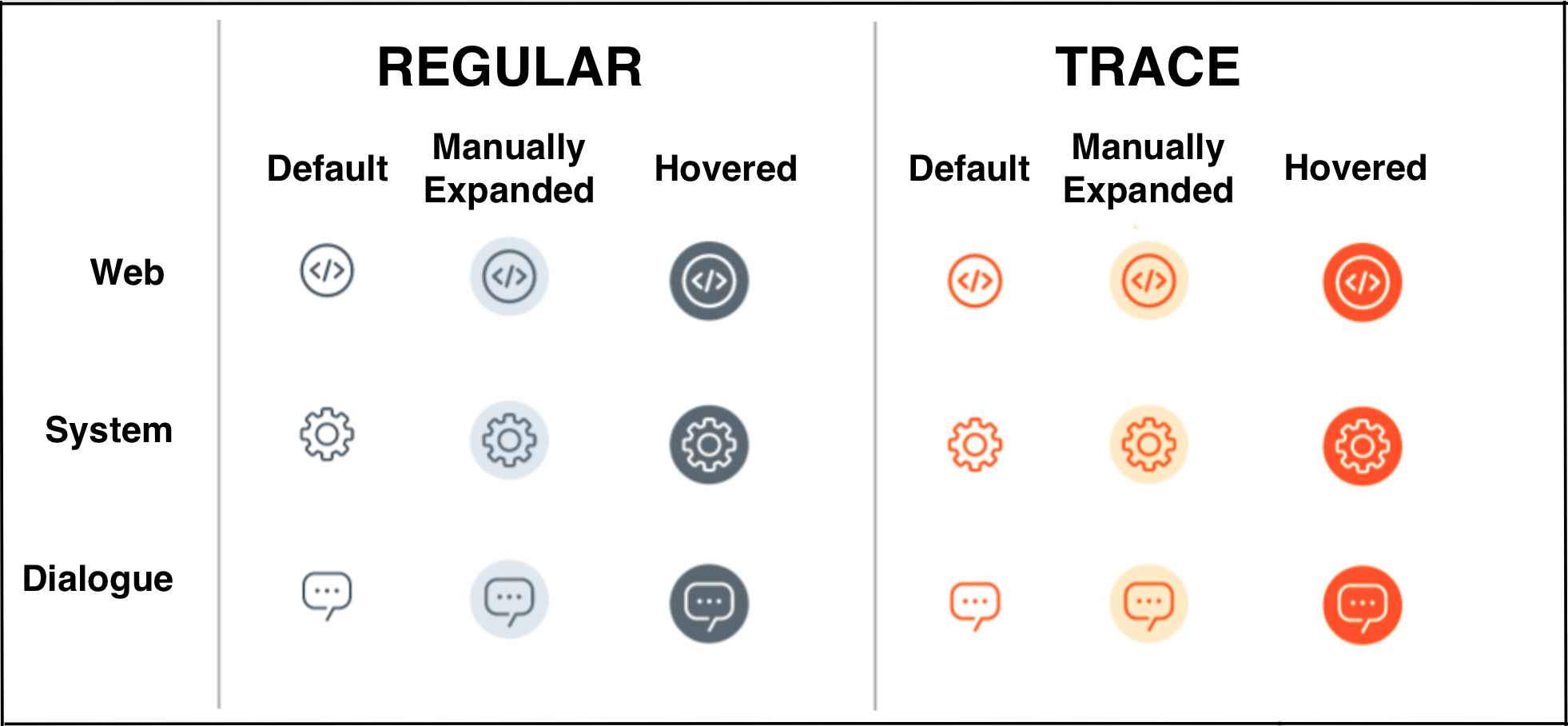}
\label{fig:node-overview}
\caption{An overview of the node icons. Trace nodes indicate the states traversed  by the actual conversation in the dialogue plan.}
\end{figure}

\subsubsection{Conversation Panel}
Conversation Panel forms the left side of the interface and contains the entire dialogue exchange that took place between the agent and the user. This dialogue was governed and led by the plan in the Dialogue Visual on the right.

\subsubsection{Dialogue Plan Visual}
The visual forms the right side of the interface and mimics the dialogue agent generated by our planner. It is represented by a directed graph where elements collectively represent intents, dialogue states, actions and goals.
\begin{itemize}
    \item Each \textbf{node} in the graph represents a \textit {state in the agent's plan}. These nodes are categorized based on the following characteristics:
    \begin{itemize}
        \item[--] \textbf{scope}: state-context set by the outcome of the action execution leading to the node. The state is represented by a \textit {dialogue} node if the intent of the state is to send out a dialogue to the user, a  \textit {system} node if an the state executes an auxiliary action or a \textit {web} node if the state executes a web service.
        \item[--] \textbf{type}: the stage of the state in the dialogue plan. A state representing the start of the conversation is represented by the \textit {root} node and the state marking the resolution of all the established intent(s) is represented by the \textit {goal} node. Every other intermediate state is represented a \textit {regular} node.
    \end{itemize} 
    \item Each \textbf{edge} represents an action's realization which upon execution updates the state and context of the variables in the plan. When an action is executed, the corresponding edge becomes a part of the larger path, \textit {Trace}, leading from the root node to the goal node. All other edges remain as unexecuted within the plan.
    \end{itemize}

The intention of the interface is to assist the designer in viewing the \textit{Trace} as it unfolds under the guidance of the agent's plan. The nodes(states the system finds itself in) and edges(realizations of the actions executed) that form the \textit{Trace} are highlighted with orange color as opposed to the rest of the graph to make it easier to differentiate. To achieve the larger objective, the interface connects the \textit{Conversation Panel} and the \textit{Dialogue Plan Visual} using multiple interactions.

\subsection{Interactions}
\begin{itemize}
\item \textbf{Expand/Collapse} - Facilitates viewing different parts of the dialogue plan for clarity of control flow by clicking on individual nodes to expand or collapse(if already expanded) the children of that node. Clicking also highlights the node icon.
\item \textbf{Hover} - This interaction is key to connecting both sections of the interface and it works in multiple ways :
\begin{itemize}
        \item[--]Hovering over a node highlights that node and all of its outgoing edges.
        \item[--]Hovering on an edge only highlights the edge.
        \item[--]The blue pane at the bottom of the graph(Figure 7) displays information about the graph element being hovered.
        \item[--]If the hovered element is mapped to message(s) in the conversation panel, the corresponding message gets highlighted as well.
        \item[--]Hovering on a message in the conversation panel highlights the corresponding graph element as well.
\end{itemize}
\item \textbf{Undo/Redo} - Allows the designer to go back and forth between dialogue plan visual changes at any point in time. 
\item \textbf{Expand all/Collapse all} - View the entire dialogue plan at once or go back to the \textit{root} node for diagnostic purposes.
\item \textbf{Actual path} - Displays only the parts associated with \textit{trace} and moves the slider to max step value. Additionally, children of all the trace nodes are displayed as well but are greyed out. 
\item \textbf{Slider} - Conversation between a user and the agent goes through various steps to conclude itself. Our framework records the state of the agent's plan and the dialogue at each step. The designer can play out these steps sequentially or choose to investigate a particular step by moving the slider. Clicking on \textit {Actual Path} takes the slider to its maximum value (i.e., most recent state of conversation). Although, after this action, the designer needs to reset the slider manually if viewing the complete trace is no longer desirable.

\item \textbf{Reset} - Re-centers the graph in the window to recover from manual panning/zooming that might have caused the dialogue plan visual to become off-centered.
\end{itemize}

\subsection{Modes}
Two primary modes are presented to the designer for interacting and diagnosing the generated dialogue plan and the conversation. Each mode caters to a different and potentially insightful requirement of the dialogue designer.

\subsubsection{Default Mode}
In this mode the dialogue designer engages with the interface pretending to be an actual user. The conversation window is open for initiation by the designer and the agent plan is pre-loaded rendering the visual to just display the root node for now. The dialogue started by the designer expands and updates the trace as well as the visual based on the pre-loaded agent plan in real-time. This mode helps the designer to preview how the agent plan would perform when deployed and use this information to resolve any glitches/unwanted segues encountered during the interaction.

\subsubsection{Replay an Existing Log}
This mode focuses on the primary diagnostic need of the designer to play out an existing user-agent conversation. The existing dialogue is displayed in the conversation panel along with the corresponding agent plan in the visual. In this mode, the designer can view how the conversation panned out under the existing agent plan, and use this information to modify the current specification to resolve an egregious conversation.

% \subsection{Discussion}
% Chatbot technology has seen increased interest in recent years from a variety of fields including games\footnote{\url{http://www.chatmapper.com/}}\footnote{\url{https://www.demigiant.com/plugins/outspoken/index.php}}, customer service\footnote{\url{https://botmock.com}}, and general business-to-business applications\footnote{\url{https://www.ibm.com/watson/ai-assistant/}}\footnote{\url{https://dialogflow.com/}}\footnote{\url{https://azure.microsoft.com/en-us/services/bot-service/}}. Our tool is uniquely situated in the analysis of generated dialogue agents (as opposed to the creation of them through tree-like interfaces). In contrast with existing services, we aim to visualize the overlay of chat logs on large and complex dialogue agents, and in particular focus on the \textit{flow} of conversations through the agent's plan.
% 
% Next steps for the development of the interface include (1) showing an aggregate view of the conversations with the agent; and (2) the ability to rollback the conversation permanently and explore another path in the dialogue.

%% file: sections/evaluation.tex
\section{Evaluation}
\label{sec:eval}

% \begin{itemize}
%     \item Model -vs- Plan size evals
%     \item (\textit{maybe}) Human-study evals
% \end{itemize}

We began with motivating the promise of declarative design 
in being able to capture complex processes with exponentially 
smaller size of specification.
We provide a empirical evidence of this in Section \ref{subsec:scaleup}.
As noted in Section \ref{ssec:planning}, and later in Section \ref{ssec:state-context-mapping}, retaining the complex nested \textit{and-or} structure of the effect trees has several advantages:
we demonstrate this with an experiment on simulated data in Section \ref{subsec:exe}.

\subsection{Scale-up of Declarative Representation}
\label{subsec:scaleup}

\begin{table}[ht]
\centering
\begin{tabular}{r|r|c|c|c|c}
\midrule
\multicolumn{6}{c}{\bf Car Inspection Domain} \\
\midrule
& number of parts & 1 & 2 & 3 & 4 \\
\midrule
\multirow{2}{*}{size of specification} & number of variables & 5 & 6 & 7 & 8 \\
& number of actions & 4 & 5 & 6 & 7 \\
\midrule
\multirow{2}{*}{size of graph} & number of nodes & 7 & 15 & 31 & 63 \\
& number of edges & 14 & 44 & 114 & 272 \\
\midrule
& time to generate (secs) & 0.005 & 0.009 & 0.014 & 0.023 \\
\bottomrule
\end{tabular}
\begin{tabular}{r|r|c|c|c}
\midrule
\multicolumn{5}{c}{\bf Credit Card Recommendation Domain} \\
\midrule
& number of cards & 1 & 2 & 3 \\
\midrule
\multirow{2}{*}{size of specification} & number of variables & 24 & 26 & 28\\
& number of actions & 15  & 16 & 17\\
\midrule
\multirow{2}{*}{size of graph} & number of nodes & 36 & 58 & 482\\
& number of edges & 127 & 207 & 1857\\
\midrule
& time to generate (secs) & 0.037 & 0.135 & 0.963\\
\bottomrule
\end{tabular}
\caption{Exponential scale-up from declarative specification.}
\label{tab:scale-up}
\end{table}

To demonstrate the scale-up from the size of the declarative 
specification to the size of the generated dialogue graph,
in Table~\ref{tab:scale-up}, 
we take two example domains and report on the 
size of the specification (in terms of the number of 
actions and variables in it) and the size of the 
composed controller (in terms of the number of 
nodes and edges in it).

\begin{itemize}
\item[-] {\bf Car Inspection Domain:} 
We introduced this domain earlier in Figure~\ref{fig:car-inspection-log}.
It involves inspecting various parts of a car with the help of an
assistant through dialogue.
In Table~\ref{tab:scale-up}, we vary the number of parts 
and show the complexity of the composed agent. 
The dialogue graph illustrated in Figure~\ref{fig:generated-plan-inspection}
corresponds to the specification with all four parts.
\item[-] {\bf Credit Card Recommendation Domain:} 
This domain helps a customer navigate different choices of 
credit cards in terms of what features they offer (such as transaction
fees, low intro APR, annual fees, etc.) -- the agent 
elicits preferences from the customer and boils down
the choices to the most relevant card.
In Table~\ref{tab:scale-up}, we vary the number of cards available 
on top of the basic preference elicitation process so as 
to illustrate the scale-up with every added card. 
\end{itemize}

Clearly, from the size of the composed processes, 
the resulting complexity of the dialogue agents is far beyond
reasonable expectation to design manually, as 
is the prevailing method in the current suite of public offerings for dialogue design.
Furthermore, as is evident from the table,
the planner very efficient in computing the solutions, which also helps
considerably in the iterative design and debugging process.

Another interesting observation in Table~\ref{tab:scale-up} is
that a big chunk of the size of the specification corresponds
to the core process captured in it -- i.e. subsequent 
additions only add 1 or 2 variables and actions.
This illustrates a powerful feature of the paradigm of
declarative design, in that the designer 
is empowered to be able to make small changes or updates to the specification 
and affect large portions of the composed process.
This is crucial to maintaining agents of this sophistication,
as the underlying processes change and evolve over time.

\subsection{Execution Framework}
\label{subsec:exe}
To demonstrate the effectiveness of our execution approach, we focus on the time it takes to do the determination. This speaks directly to a core novelty of our work: the execution of actions with complex nested effects.
Recall that the Algorithm \ref{alg:parnes} (which we called \textit{parallel nested algorithm}) evaluated the \textit{and} nodes in parallel and it used the nested structure of the graph so it recursively evaluated only one sub-tree of an \textit{oneof} node. We compared it with three naive alternatives:

\begin{enumerate}
    \item \textit{parallel flat}: does not care about the tree structure and evaluates all the determiners in parallel and then evaluates the outcome (which takes negligible time compared to the determiners)
    \item \textit{sequential nested}: this is the same as our proposed approach except that the \textit{and} nodes are not evaluated in parallel but sequentially (representing the effect a single core machine would have on computation)
    \item \textit{sequential flat}: the same as \textit{parallel flat} but evaluates the determiners sequentially
\end{enumerate}

\begin{figure}[p]
    \centering
    \includegraphics[width=0.9\linewidth]{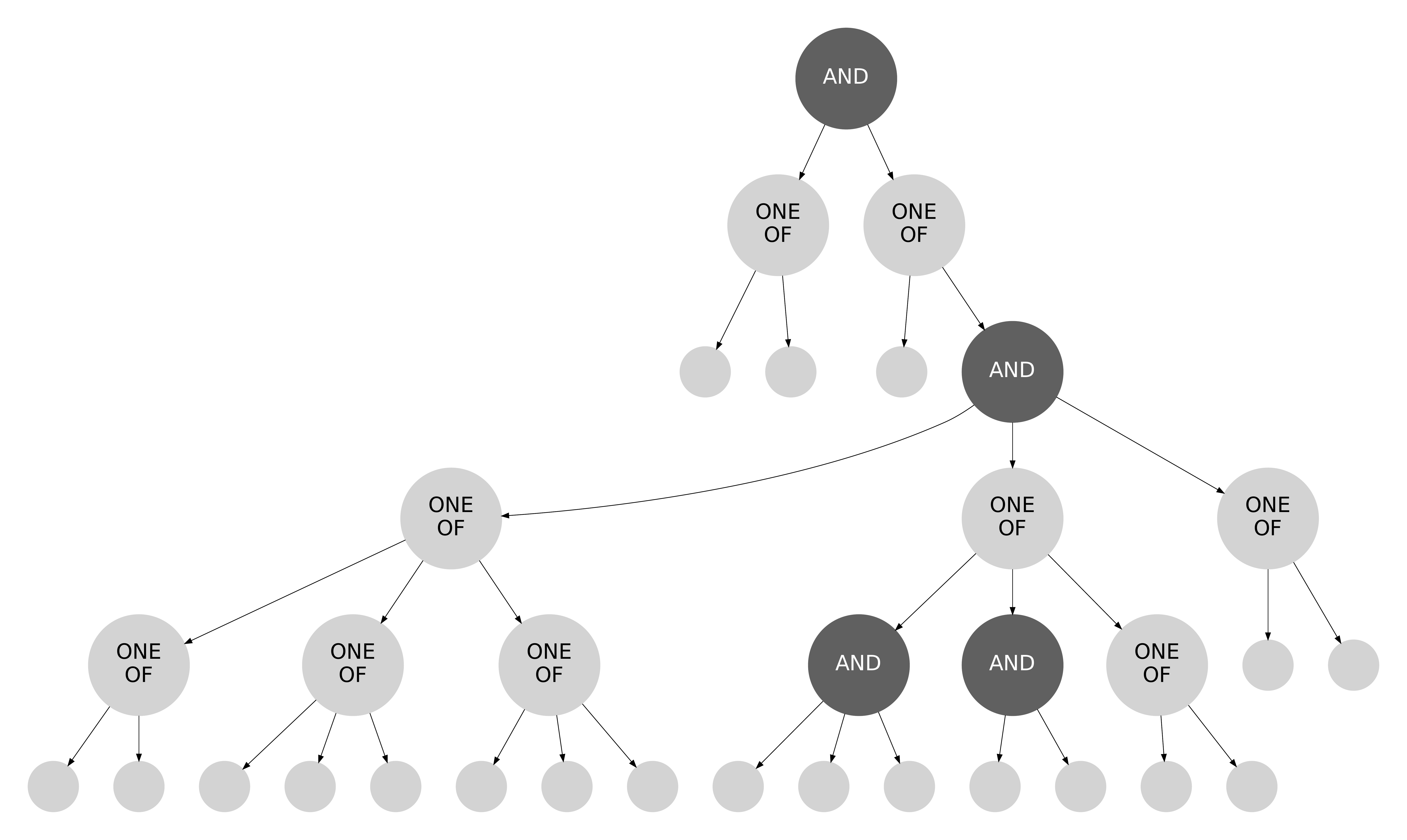}
    \caption{General complex effect tree, which is the most interesting from the practical application point of view. It could be, for example, a slightly more advanced version of the graph in Figure \ref{fig:effect-example}.}
    \label{fig:effect_graph_0}
\end{figure}

\begin{figure}[p]
    \centering
    \includegraphics[width=0.9\linewidth]{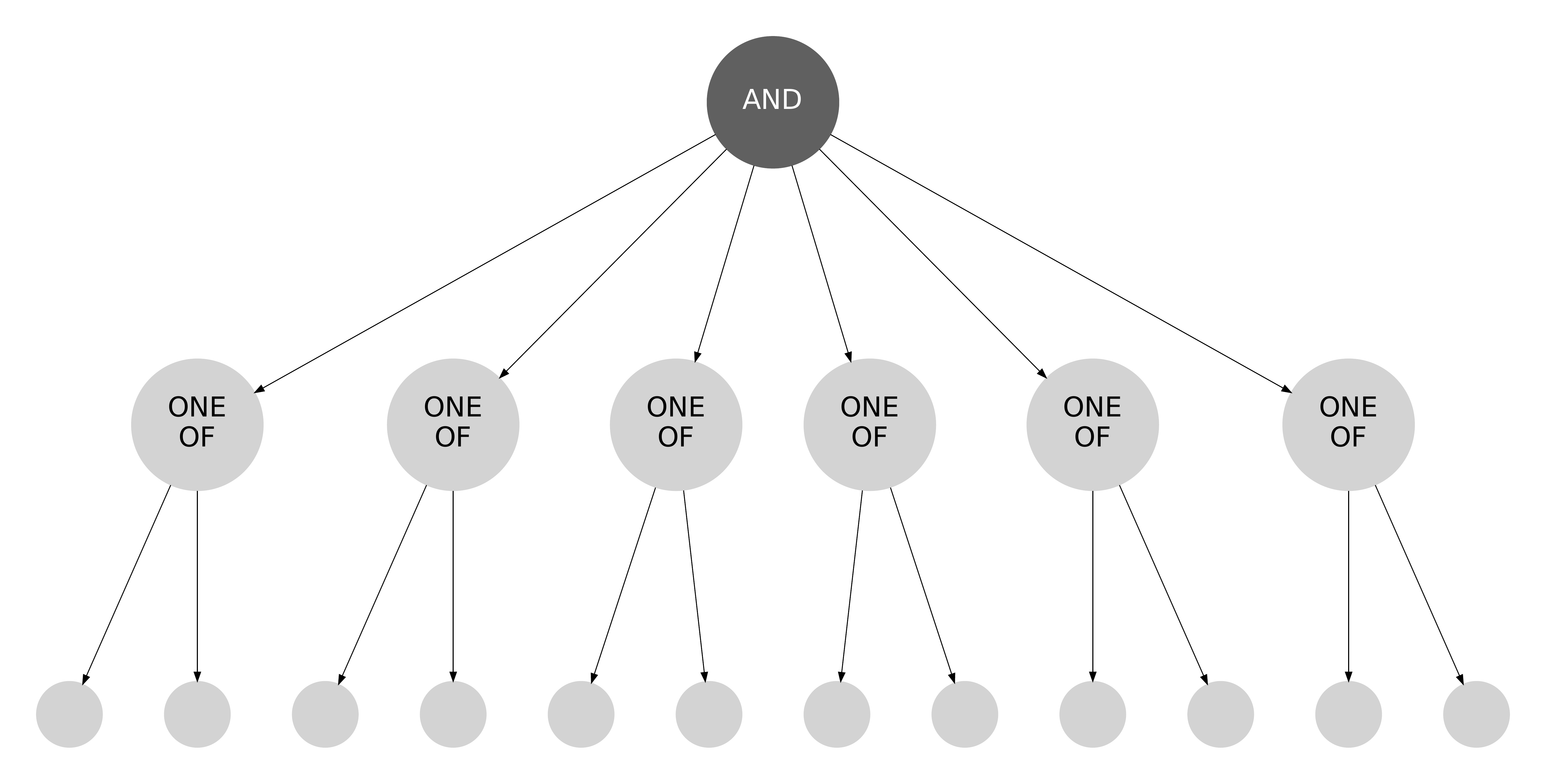}
    \caption{Flat effect tree that serves as a benchmark for how well the algorithms scale with growing number of determiners, that could be run in parallel.}
    \label{fig:effect_graph_1}
\end{figure}

\begin{figure}[p]
    \centering
    \includegraphics[width=0.9\linewidth]{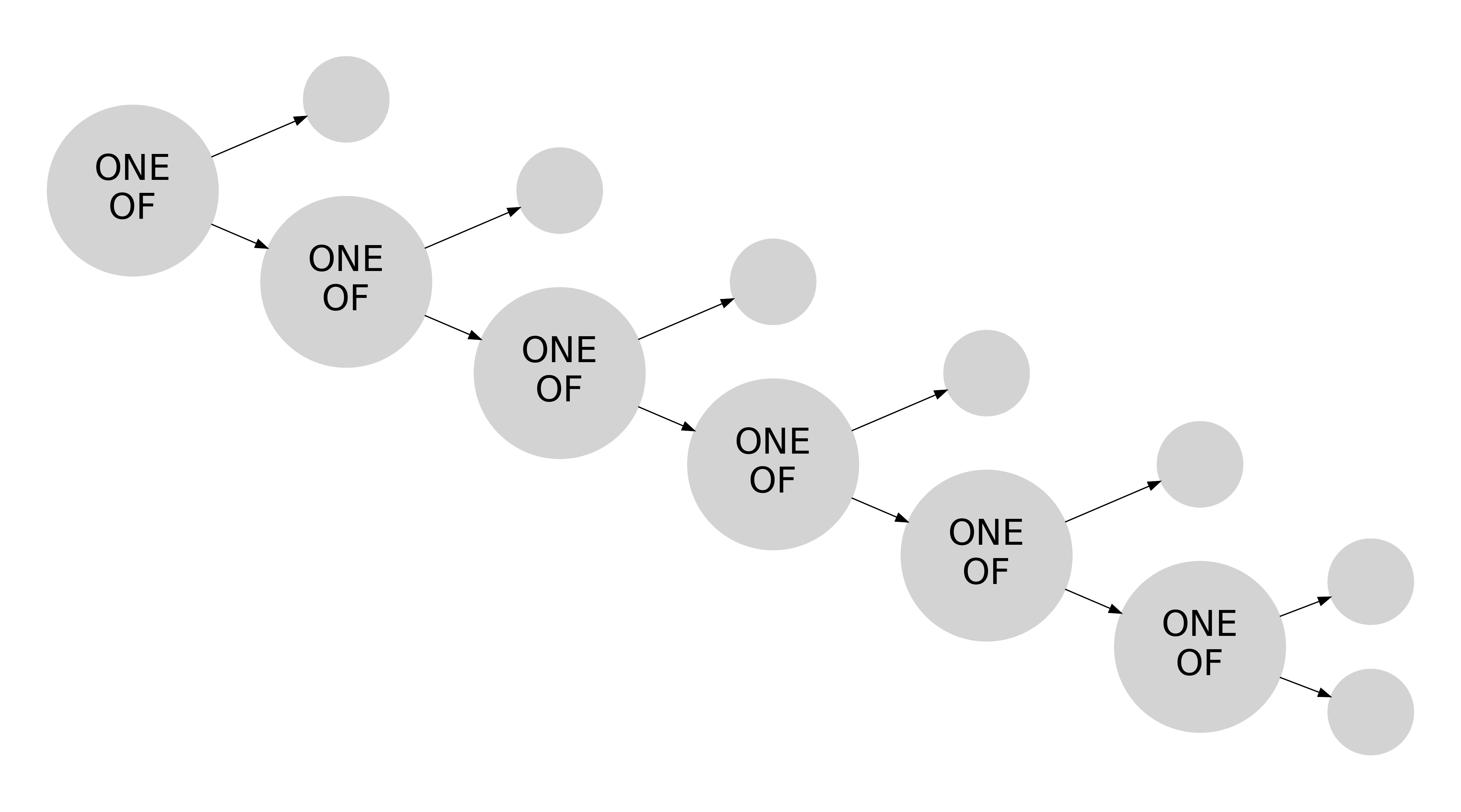}
    \caption{Deeply nested effect tree that serves as a benchmark for how well the algorithms scale with growing number of nested determiners.}
    \label{fig:effect_graph_2}
\end{figure}

We simulated the time to evaluate the outcome of three representative effect trees, which are depicted in Figures \ref{fig:effect_graph_0}, \ref{fig:effect_graph_1}, and \ref{fig:effect_graph_2}. The size and the structure of the trees remained exactly the same during the simulation and only the outcomes and times needed to run the determiners changed.

\begin{figure}[p]
    \centering
    \includegraphics[width=0.9\linewidth]{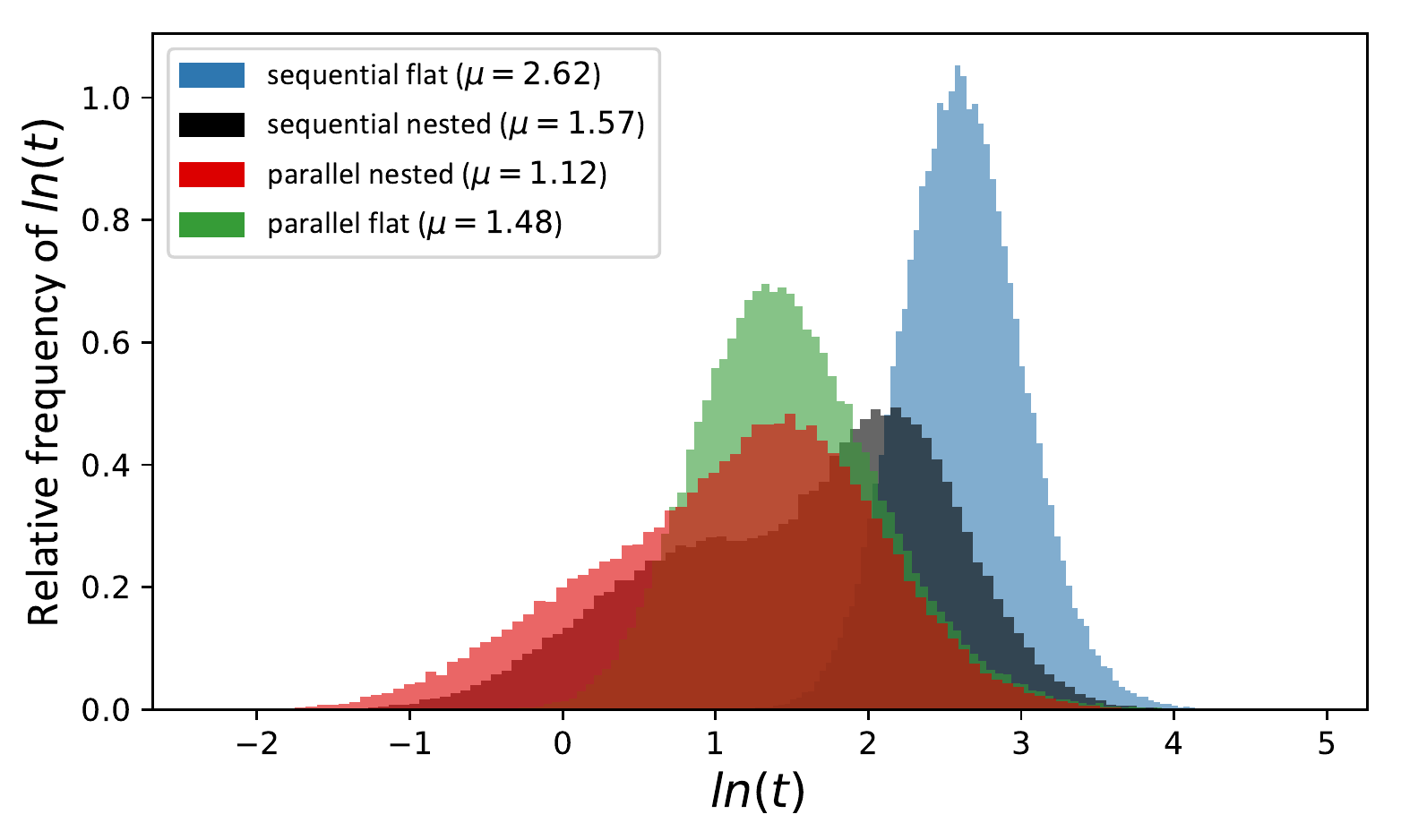}
    \caption{Distributions of logarithm of time to determine outcome of effect tree in Figure \ref{fig:effect_graph_0}.}
    \label{fig:determination_time_distributions_0}
\end{figure}

\begin{figure}[p]
    \centering
    \includegraphics[width=0.9\linewidth]{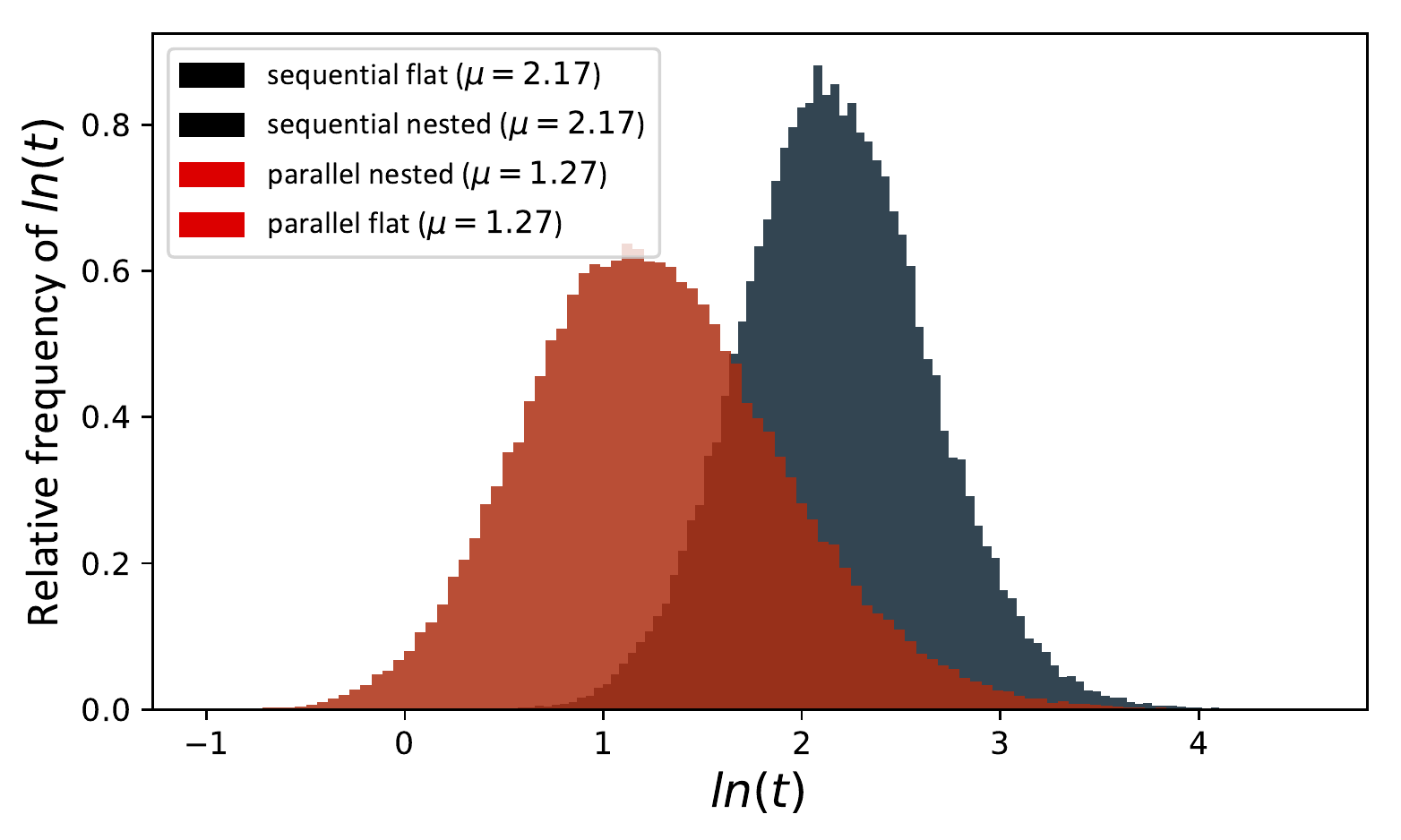}
    \caption{Distributions of logarithm of time to determine outcome of effect tree in Figure \ref{fig:effect_graph_1}.}
    \label{fig:determination_time_distributions_1}
\end{figure}

\begin{figure}[p]
    \centering
    \includegraphics[width=0.9\linewidth]{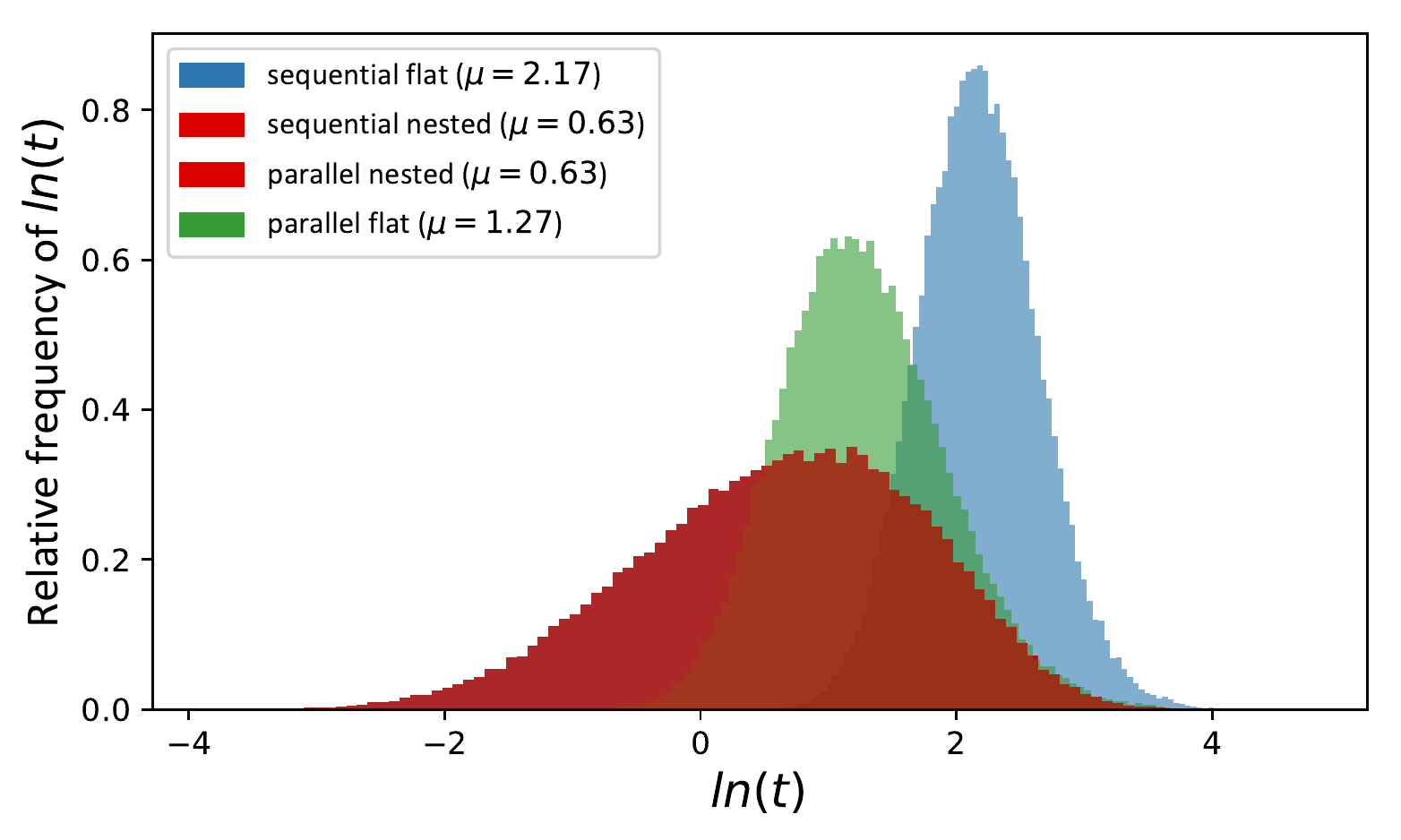}
    \caption{Distributions of logarithm of time to determine outcome of effect tree in Figure \ref{fig:effect_graph_2}.}
    \label{fig:determination_time_distributions_2}
\end{figure}

Let us assume in this section, that all the determiners can run in an arbitrary order. If running some of the determiners required other determiners to run before, we could not use the flat algorithms. In this case, running a determiner in a node requires all of the determiners in its parent nodes to run prior, and the nested algorithms are the natural way to evaluate the tree. 

We restrict our attention only to the time it takes to evaluate the determiners in \textit{oneof} nodes since this is the most expensive operation. We assume these times are all independent following \textit{LogNormal(0,1)} distribution. This was chosen, as it is a reasonable distribution for the time to make an API call in our experience with the deployed system. To gain a better perspective the actual running time of one determiner can be on the order of milliseconds to seconds. 
We also need to simulate the different outcomes. We do this recursively going top down in the tree selecting nodes that will belong to the outcome. The only decision we have to make in the process is, what child of a previously selected \textit{oneof} node to select next. We do this independently on selections in the previous \textit{oneof} nodes so that all the child nodes have equal probability of being selected.

We run 100,000 simulations, sampling the determiner execution times and the outcome at each step. Histograms of the logarithm of simulated times for the different algorithms are depicted in Figures \ref{fig:determination_time_distributions_0}, \ref{fig:determination_time_distributions_1}, and \ref{fig:determination_time_distributions_2}. We chose to display histograms of $ln(t)$ instead of the actual times for clarity when comparing them with the histogram of log-times to run a single determiner, which would be depicted just as a probability density function of a standardized normal distribution. 

Obviously we would expect the parallel versions to be faster than their sequential counterparts. The interesting comparison is between the \textit{parallel nested} and \textit{parallel flat} approaches. It is clear that the \textit{parallel flat} will have to call on average more determiners, so it is more computationally expensive, but it is limited only by the time it takes to run the slowest determiner in the tree whereas the \textit{parallel nested} algorithm waits for determiner in an \textit{oneof} node to finish before running any determiner in its sub-trees. However, as we can see from the histograms, the \textit{parallel nested} approach still outperforms \textit{parallel flat} on average since it often does not have to call the slowest determiner at all. This is perhaps quite surprising considering all the determiner times were drawn from independent identical distributions. 

In each step of the simulation, we used the same outcome and sample of determiner times for all the four algorithms, so it is not a surprise that the histograms of both parallel and both sequential methods coincide in Figure \ref{fig:determination_time_distributions_1}, since the \textit{parallel flat} and \textit{parallel nested} algorithms are the same for the graph in Figure \ref{fig:effect_graph_1} as are their sequential counterparts. This is due to the fact that there is no nested structure in the \textit{oneof} nodes which could be exploited by the nested algorithms. The same thing happens for \textit{sequential nested} and \textit{parallel nested} histograms in Figure \ref{fig:determination_time_distributions_2}. This is because there are no \textit{and} nodes in the graph, so there is no chance for the algorithms to run them in parallel. 

Generally, we find that following the complex effect structure for determination, and using parallel execution whenever possible, represents a clear advantage.

%% file: sections/related.tex
\section{Related Work}
\label{sec:related}

\subsection{Declarative Dialogue}
Research of planning techniques in dialogue is usually limited to developing formal frameworks that describe subsets of dialogue phenomena.
However, the necessary steps for making the frameworks understandable by planning non-experts, such as dialogue designers, allowing mass adoption of the techniques, 
are often neglected.

Work of~\cite{steedman2007planning} describes dialogue by declarative actions expressed as formulas from formal logic grounded around a set of axioms.
The authors present the formalism on mixed initiative dialogues that can be described only by initial facts of the parties and communication rules.
A similar approach can be seen in the work of \cite{black2014automated} where planning is used to optimize the likelihood of persuading an opponent that a set of given facts holds.
During the dialogue, the persuader asks the opponent for its belief so the optimal facts for presentation can be chosen.
The dialogues are generated only on an abstract layer skipping challenges of using natural language.

Complexity of generating real natural utterances is demonstrated in work of~\cite{garoufi2010automated}.
Their system uses planning techniques to provide navigation information in a known environment.
The system optimizes the instructions in a way which enables their transcription into natural sentences.
The authors present an example when the bot generates a movement instruction first to be able to use object referencing in the next turn.

The interesting work of~\cite{petrick2013planning} shows a dialogue system used in real world conditions.
Their physical bot uses planner to combine multiple sensory inputs together with dialogue control in bartender domain.
The bot deals with frequent environment changes (e.g. new customer came) by recalculating the plan. This work, however, is far more domain specific (to the bartender setting) than we set out to capture with arbitrary dialogue design.

Planning supports multi-modality also in work of~\cite{romero2017cognitive}.
They use complex preconditions consisting of multi-modal features like user rapport level estimation, non-verbal strategies (like smile) etc.
The features allow the system to drive the dialog more naturally which results in measurable increase in user rapport level.

% Much of what we put into the related list on the mega plan4dial paper. Can be found in the resources/plan4dial-paper-related.tex file.
%
% other considered references:
% \cite{srivastava2018chatbots} -- nothing about declarative dialogue
% Mathur, V., \& Singh, A. (2018). The Rapidly Changing Landscape of Conversational Agents, 1–14. Retrieved from http://arxiv.org/abs/1803.08419 --nothing about declarative dialogue

\subsection{Execution Monitoring}
Much execution monitoring work focuses on the theoretical properties of the system without diving into the details of how the physical system should map to the planner's abstraction of the environment, or forgo entirely the notion of a framework for determining action outcomes.
\textsc{TpopExec} \cite{DBLP:conf/ijcai/MuiseBM13} addresses the theoretical properties of deterministic temporal partial order plans (as opposed to the conditional plans we consider).
The Kirk and Drake systems \cite{block2006robust,DBLP:journals/jair/ConradW11} similarly focus on partial order temporal plans, but with choice points in the execution (akin to contingent solutions). A similar approach is employed by the Razor system for compiling contingent plans in an information gathering setting \cite{DBLP:conf/ijcai/FriedmanW97}. However, these works mainly deal with compiling one form of plan with unrealized choice points, to an executable or \textit{dispatchable} form: the complexity of non-determinism is trivial (compared to the full nesting of \textit{and} and \textit{oneof} that we consider in Section \ref{ssec:complex_outcome_determination}), and there is no focus on the mapping to realized systems (i.e., correspondence between context and state, and separation of action execution and realization). Other work extends the notion of uncertainty to temporal durations \cite{DBLP:conf/aips/KarpasLYW15}, but again this is mainly theoretical in nature and focuses on a separate set of challenges.

Closer to our executor strategy, the \textsc{IxTeT-eXeC} system \cite{DBLP:conf/aaai/LemaiI04} focuses on some of the challenges an executor faces when plans are deployed; the key difference being that their focus is on temporal actions and not contingent plans with modeled uncertainty (i.e., complementary aspects of execution). The system is built on the OpenPRS procedural executor \cite{openprs}, and unexpected failures to the execution are handled through replanning and plan repair approaches.
Languages for executors have also been proposed, such as PLEXIL \cite{verma2005plan} and RMPL \cite{DBLP:journals/pieee/WilliamsICE03}. Similar to the work cited above, the focus of these languages is to place temporal-based plans in a dispatchable form, with the focus on adhering to the semantics at the planning level of abstraction and temporal consistency of the execution.

The languages of PLEXIL and Esterel \cite{DBLP:conf/concur/BerryC84} are programming languages for autonomous systems. Their relation to our work is in the output format of the contingent plans we produce, but not in the higher-level philosophy of declarative modeling. Of the two languages, PLEXIL uses a representation closer to the complex nested effects we describe (i.e., a graph of nodes with key similar interpretation). Both PLEXIL and Esterel are well-defined programming languages with rich expressibility. Our work aims at addressing a different set of challenges with respect to execution: namely the specification and handling of complex action effects, and the challenges / opportunities surrounding the connection between state and context.

In terms of the state and context mapping that we propose, whenever the context $C$ is defined for a fluent $f$ (i.e., $C(f) \neq \bot$), then we can view $f$ as representing the fact that we know the value of a particular variable with a rich domain. This mirrors the idea of the $Knows$ predicate in \cite{DBLP:conf/aaai/ScherlL93} and $\mathit{KnowIf}$ ($KIF$) variables in \cite{DBLP:journals/aamas/BrennerN09}. Also related is the recent work on integrating PDDL plan execution with the CLIPS rule-based production system \cite{niemueller-intex18}. In this work, various models are defined, including the planner model (which corresponds to our planning state view) and world model (which corresponds loosely to our context view). Key differences, however, include our focus on non-deterministic settings and the direct relation to implementations of action execution and outcome determination.

The connection between high-level action specifications and low-level sensor / behaviour modalities is described in the work focusing on Object-Action Complexes \cite{DBLP:journals/ras/KrugerGPPSWUAKOAD11}. The connection to our work is in the correspondence between planning actions and the action execution functions we use to realize them. However, there is little more to be drawn from the parallel, aside from the proposed approach of maintaining two views on the world.

Similar approaches can be seen in the robotics community through works such as the KNOWROB, SkiROS, and ROSPlan systems \cite{DBLP:conf/iros/TenorthB09,rovida2017skiros,DBLP:conf/aips/CashmoreFLMRCPH15}. The robotics focused systems similarly solve the task of linking the planning state and the execution context. Differences include their focus on temporal execution and the monolithic view of action execution / determination. Perhaps the most mature of the existing work, ROSPlan, accepts a variety of input plan specification languages, including simple sequential plans, the Esterel plan language (which represents a temporal plan without the uncertain contingencies we aim to support), contingent plans represented as state-to-action policies, and Petri Net Plans \cite{DBLP:conf/atal/ZiparoINPC08}. Both the contingent plan and Petri Net Plan representations fail to capture much of the sophistication we introduce in this paper, including the separation of action execution / determination, and the nested functionality of non-deterministic effects. Conversely, aspects such as temporal action execution, loops, and interrupts are components that the various ROSPlan interfaces are capable of expressing that we forgo in our present work.

Similar to the discussion in Section \ref{sec:exec} on embedding our work into a larger EM framework, there is potential for us to extend a system such as ROSPlan to capture our methodology for the robotic setting. The advantage would largely come in the form of the improved expressivity in action effects for the declarative specification of robot behaviour.

%% file: sections/discussion.tex
\section{Summary}
\label{sec:summary}

We have presented an end-to-end solution for the creation of goal-oriented multi-turn dialogue agents. At its core, automated planning technology for handling non-determinism is used to create large and complex dialogue agents from compact declarative specifications. Beyond describing the planning abstraction for the setting, we also further explored the model acquisition process for dialogue designers and the execution engine required for deploying the agents. We have implemented all aspects of our system and demonstrated its effectiveness empirically.

We anticipate that this modeling paradigm will not only reduce the effort in building robust dialogue agents for enterprise use, but also make the design process of such bots accessible to a wider range of users (and thus to a wider range of uses). Initial steps are already under way to assess which aspects of this work can be integrated into IBM's Watson Assistant software platform, and we are excited about the further research potential it offers. We dive further into these ideas for future research in the next section.

\subsection{Future Directions}
\label{sec:future}

We conclude with a discussion of the {\em beta} features
highlighted in Figure \ref{fig:d3wa-inspection}.
These are specifically meant to give the dialogue designer different
pathways to tap into the core specification, especially those
who are not familiar enough with the declarative paradigm
to build it from scratch.

\subsubsection{Bootstrapping Specifications from Conversation Logs}
\label{subsec:bootstrap}

\begin{figure}[!t]
\centering
\includegraphics[width=\textwidth]{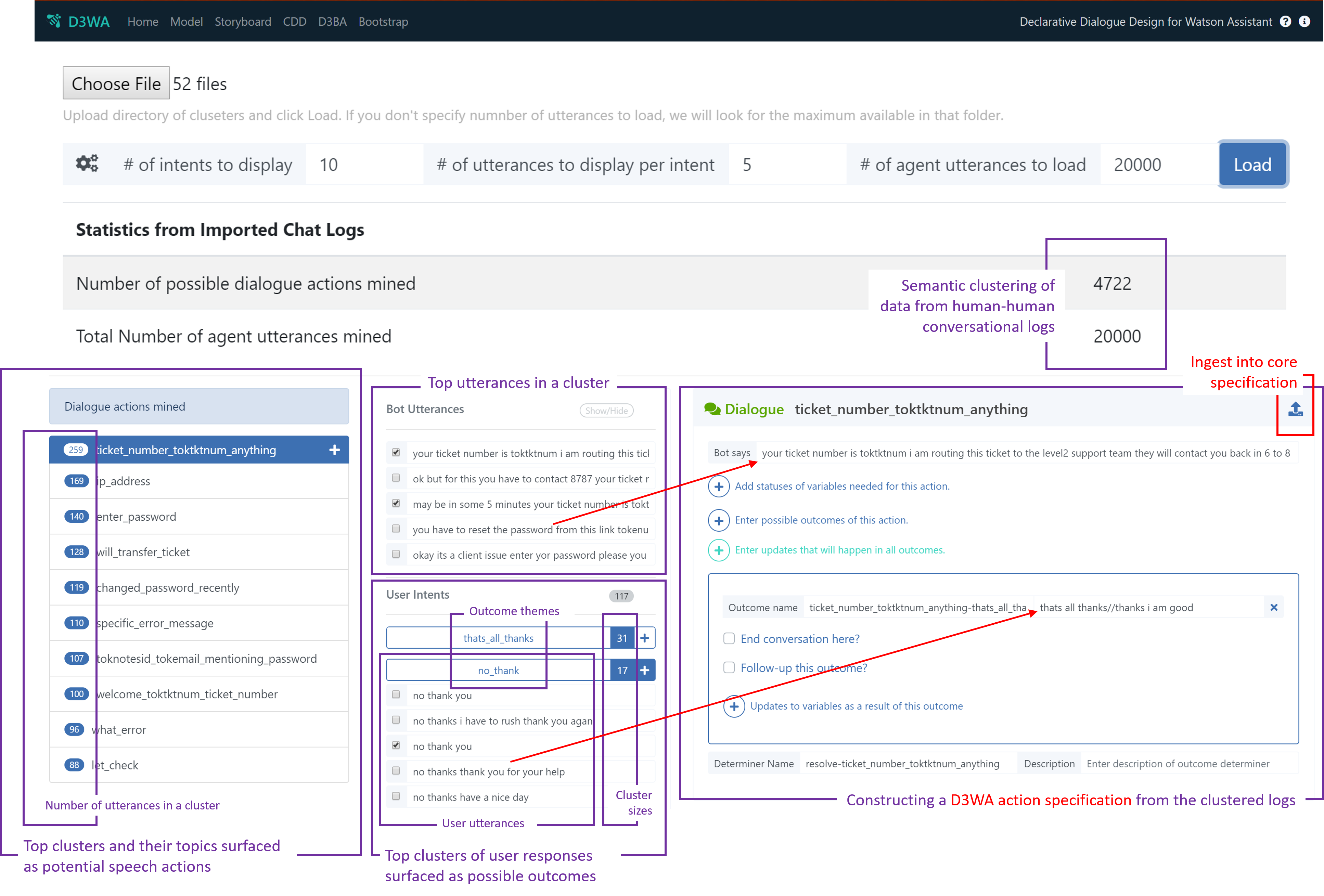}
\caption{Bootstrapping \mai\ specification from logs of conversations.}
\label{fig:bootstrap}
\end{figure}

It is often the case that a company wishing to create a dialogue agent will already have a large corpus of human-to-human conversation logs they would like to capture. To take advantage of this rich data source, we are exploring the option of \textit{bootstrapping} the declarative model.

Illustrated as part of our interface in Figure \ref{fig:bootstrap}, the process is roughly as follows:

\begin{enumerate}
    \item Tokenize the entities in all of the logs.
    \item Semantically partition all of the human-agent utterances, and surface those partitions as candidate actions.
    \item Semantically partition all of the one-step end-user responses, and surface those partitions as candidate outcomes.
    \item Prepopulate an action description with the appropriate agent utterance examples, and training examples for the outcomes.
\end{enumerate}

The semantic parsing is done using off-the-shelf seq-to-seq models that provide a latent embedding for an utterance, and partitioning is done based on these embeddings. Note that while we extract candidate actions and their outcomes (along with example utterances), we do not extract any aspects of the world model. Nonetheless, this general method of bootstrapping is an exciting new area for human-in-the-loop induction of action models in the dialogue agent setting.

\subsubsection{\cdd\ -- Constraint Driven Dialogue}

One of the challenges in shifting to a declarative paradigm
is the shift in mental model of the end user of the system, i.e.
the dialogue designer, who is accustomed to an imperative style of implementing the dialogue agent.
Keeping this in mind, in Section \ref{subsec:sugar}, 
we introduced some modeling enhancements to 
make this journey easier.
In the \cdd\ tab, we go one step further and provide 
an even simpler abstraction of declarative design --
here, in ``constraint driven dialogue specification'',
the designer provides just the goals, i.e. the 
different ways the conversation can end, and the planner
figures out all of the different ways the capabilities of the 
bot can be combined to get there. 
In effect, this becomes a more advanced 
form of slot-filling technique.
It is much simpler to specify than the full declarative model, but it comes at the cost of losing some control over the underlying process.
Our interface does allow compiling this specification 
back into \mai\ format so that much finer details can 
be included beyond this initial starting point.

\subsubsection{Storyboarding}

While in Section ~\ref{subsec:bootstrap}, we provided an interface to
derive insights from large sets of conversational logs together, 
in {\em Storyboarding} we allow the designer to investigate 
a single conversation in order to:
1) validate against the current specification 
-- i.e. ask questions like: {\em is this (part of) a valid conversation according
to the specification?} or {\em does this conversation achieve the goal?}; and 
2) tag parts of the sample conversation with their associated actions and outcomes (to ask more pointed forms of the questions above) to act as a spring board for 
the rest of the specification. 
This is thus yet another pathway for the dialogue designer to 
feed into the core tab.
The system answers questions of these forms using the {Meta-Writer}
assist feature described next.

\subsubsection{The Meta-Writer Assist}

A common issue that arises at the time of design,
is the question of unsolvability -- {\em what happens when
a designer writes a domain that has no solution?} 
The ``meta-writer'' feature tries to assist the designer in such scenarios.
In order to do this, the meta-writer internally casts the domain writing 
process as a separate planning problem.
This meta-problem models the minimum requirements of a valid model
and monitors for fulfillment of those basic requirements as the designer is constructing the bot.
If there are outstanding items to be modeled, it suggests possible dialogue actions
to the user who can directly add them into the model of the agent.
The meta-writer uses \cite{ramirez2009plan} to compile the designer's actions on \mai\ into observations that can be compiled into a meta-planning problem.
It uses FAST-DOWNWARD \cite{helmert2006fast} as the underlying planner
to solve this.
At this time, the suggestions to fix the specification are fairly simplistic 
(e.g. adding a new action, outcome, or update which may have nothing to do 
with the intent of the designer) and gives no indication of what is actually wrong
with the specification that failed to produce a solution.
In such cases, instead of just surfacing the alert and a possible fix,
we are currently exploring adding more diagnostic
information in the form of explanations explored
in recent literature \cite{sarath-unsolvability}.

\subsubsection{\dba\ -- Optimization of Business Processes}

Finally, at the core of declarative design of dialogue agents
is the specification of {\em process}. 
This can be any process in general, not particularly 
having anything to do with conversational interfaces. 
We are thus currently exploring an extension of \mai\ in the context of 
optimization of business processes by means of external skills \cite{d3ba}. 
These skill specifications are imported from a catalog,
and composed with a business process in \mai\ in order to
arrive at an optimized process that shares 
the exponential scale-up and other features
of management and debugging made available by this work.
The focus of the actions in this setting is far more geared
towards the web- and system-based variety.

%% file: sections/acknowledgements.tex
\hide{
\acks{
Coming soon...
}
}